# A cross-process welding penetration status prediction algorithm based on unsupervised domain adaptation in laser and TIG welding

Sen Li, Haichao Cui*, Chendong Shao, Yaqi Wang, Xinhua Tang

*Shanghai Key Laboratory of Materials Laser Processing and Modification, School of Materials Science and Engineering, Shanghai Jiao Tong University, Shanghai 200240, PR China.*

**Abstract:**

Supervised deep learning has been widely used for weld penetration state classification; however, its performance often degrades significantly under domain shift, such as when transferring models between welding processes with distinct physical mechanisms—for instance, from arc-dominated tungsten inert gas (TIG) welding to keyhole-based laser welding. To overcome this limitation, we propose an unsupervised domain adaptation (UDA) framework integrated with a gradual source domain expansion (GSDE) strategy. Evaluated on dedicated TIG and laser welding datasets, our approach achieves high accuracy in both same-process and cross-process transfer tasks. Specifically, it attains average accuracies of 90.65% on TIGFH and 90.72% on LSPS in same-process settings, surpassing a supervised baseline by 35.83% and 38.87%, respectively. More notably, in cross-process scenarios, it reaches 80.48% for TIG→Laser and 81.13% for Laser→TIG, improving upon the baseline by 43.39% and 43.40%. UMAP visualizations verify that the model learns domain-invariant features while maintaining discriminative class boundaries. This method considerably lowers the relabeling cost for new welding processes and enhances the versatility of intelligent monitoring across different welding systems.

| Nomenclature | |
|---|---|
| Adam | Adaptive moment estimation |
| ADETL | Adaptive domain enhancement transfer learning |
| AMP | Automatic mixed precision |
| BPP | Beam parameter product |
| CMOS | Complementary metal-oxide-semiconductor |
| CNN | Convolutional neural networks |
| CNN-LSTM | Convolutional neural network-long short-term memory |

* Corresponding author: haichaocui@sjtu.edu.cn

| | |
|---|---|
| DNN | Deep neural networks |
| FWHM | Full width at half maximum |
| GAN | Generative adversarial network |
| GSDE | Gradual source domain expansion |
| NBF | Narrow band filter |
| LAHW | Laser-arc hybrid welding |
| LSPS | Penetration state dataset for laser welding |
| PWM | Pulse width modulation |
| RBFNN | Radial basis function neural network |
| SDA | Supervised domain adaptation |
| SGD | Stochastic gradient descent |
| ST-MDLN | Semi-supervised transfer learning-based multi-domain learning network |
| SSDA | Semi-supervised domain adaptation |
| TIG | Tungsten inert gas |
| TIGFH | Fusion hole dataset for TIG |
| UMAP | Uniform manifold approximation and projection |
| UDA | Unsupervised domain adaptation |
| VIT | Vision transformer |
| WAAM | Wire arc additive manufacturing |

# 1. Introduction

Laser welding is a key technology in advanced manufacturing. It is distinguished by its high speed, deep penetration, and narrow heat-affected zones, making it pivotal for joining diverse metallic materials like stainless steel and aluminum alloys in critical industries such as automotive, aerospace, and energy[1]. TIG welding, characterized by high arc stability, ease of control, excellent bead quality, and the absence of spatter during the welding process, has become a primary choice for single-sided welding with double-sided formation processes in sectors such as petroleum pipelines, pressure vessels, shipbuilding, and aerospace[2]. In both processes, ensuring reliable fusion of the workpiece remains a key research focus. The welding process is an inherently complex system involving the coupling of multiple physical fields (optical, thermal, and electromagnetic). Consequently, the fusion state is significantly influenced by welding process parameters. The strong interdependence between process parameters and multi-physics effects presents substantial challenges to the modeling of welding processes. Accurate modeling of the welding process is not only critical for real-time control of the fusion state but also essential for addressing the bottleneck in process parameter optimization.

In recent years, substantial advancements have been achieved in welding quality assessment by integrating sensor technology with machine learning, particularly deep learning methodologies. Early investigations predominantly concentrated on the extraction and analysis of information from individual sensors. For instance, Luo et al. [3] utilized a coaxial vision system combined with a RBFNN to estimate the geometric shapes of laser-welded small holes and predict potential welding defects. Li et al. [4] innovatively devised a LAHW penetration depth monitoring approach based on CNN, enabling precise classification of welding penetration states and effective prediction of penetration depths using front-view molten pool images. Nevertheless, these models, which rely on single-process data, often yield satisfactory results only in specific scenarios. They frequently exhibit limited generalization in complex and dynamic industrial settings. This limitation complicates the accurate detection of subtle quality variations.

Under consistent welding conditions, production environments inherently impose constrains on image quality and angles compared to controlled laboratory settings, while laboratory conditions fail to fully replicate the complexity of industrial scenarios. This results in a disparity in the feature space between datasets generated in laboratory environments and actual production data. The insufficient generalization capabilities of models lead to high-precision laboratory models being unable to meet the requirements of production environments. Consequently, the development of cross-process welding detection models faces the following critical challenges: 1) Differences in heat source and molten pool behaviors across various welding methods result in significant image heterogeneity between processes; 2) Models trained on a single process are incapable of capturing generalized multi-process features; 3) While joint training across multiple processes can improve generalization, it faces challenges such as high annotation and training costs, and incomplete scenario coverage. Therefore, it is essential to design cost-effective deep learning architectures that can integrate multi-scenario features and mitigate data distribution discrepancies. Such designs are crucial for advancing industrial intelligent welding detection systems.

To address these challenges, researchers have systematically investigated the application of transfer learning methods in the field of welding, evolving from traditional pre-trained models to more advanced semi-supervised and unsupervised approaches. Early studies predominantly focused on utilizing pre-trained deep learning models as feature extractors, enabling cross-domain knowledge transfer through fine-tuning or parameter migration. For instance, in the context of classifying weld defects in X-ray inspection, Sawsen et al. [5] investigated transfer learning-based convolutional neural network models by fine-tuning and comparing five pre-trained models, thereby demonstrating the efficacy of transfer learning in small-sample weld defect classification tasks. Similarly, for fault diagnosis caused by variations in welding thickness during robotic spot welding, Noh et al. [6] proposed a robust fault diagnosis model based on parameter transfer learning. Additionally, Pan et al. [7] introduced a transfer learning-based TL-MobileNet model, achieving superior classification accuracy. To tackle the challenges of insufficient data in arc additive manufacturing, Shin et al. [8] employed convolutional neural networks to extract image features from multiple source materials, utilizing a progressive parameter freezing strategy to incrementally learn features across various source domains, thereby facilitating spheroidization defect detection in different welding materials. Although pre-trained models have made remarkable advancements, they typically rely on large-scale labeled data. To address this limitation, subsequent studies have progressively explored semi-supervised transfer learning approaches. For example, Kumar et al. [9] proposed a novel ST-MDLN model that substantially enhanced the segmentation performance of welding defect images by leveraging knowledge transferred from extensive unlabeled data. Dai et al. [10] introduced an ADETL framework that integrates self-supervised learning and continual learning strategies, effectively utilizing incremental unlabeled data through parameter space and loss landscape analyses to improve classification accuracy for welding defect. Recently, UDA methods have attracted considerable attention due to their ability to operate without requiring labels in the target domain. The fundamental concept of UDA involves using labeled source domain data and unlabeled target domain data to learn invariant feature representations, enabling model to generalize well to the target domain [11]. Compared with conventional transfer learning methods, UDA eliminates the reliance on target domain labels, thereby significantly reducing data annotation costs and promoting broader applications in areas such as medical imaging [12], signal processing [13], and audio-video analysis [14]. For example, Zhou et al. [15] proposed a UDA-based approach to improve the generalization capability of predictive models for fusion/pore states in plasma arc welding, achieving cross-scenario predictions across different welding conditions without the need for labeled data. Furthermore, Zhang et al. [16] presented a dual-domain adaptive transfer learning strategy that not only aligns global edge distributions but also focuses on fine-grained conditional distribution alignment, incorporating multi-head self-attention mechanisms to boost the performance of feature extractors.

Although the aforementioned studies have achieved significant progress in areas such as welding defect detection and status prediction, existing UDA methods predominantly focus on addressing domain shift issues within the same welding process. These shifts are typically caused by factors such as lighting variations, camera calibration discrepancies, or noise interference. However, in practical industrial applications, the diversity of welding processes and varying parameter settings lead to substantial differences in image features, thereby hindering the generalization of models to new welding processes. To tackle this challenge, this study introduces a novel cross-process UDA method aimed at bridging the feature gap between different welding processes, enabling penetration state prediction without requiring target domain labels. To support UDA research across welding processes, this paper constructs a specialized dataset that includes laser welding and TIG welding images captured by various industrial cameras and processed with different filtering schemes. To comprehensively evaluate the effectiveness of the proposed method, we conducted model visualization, ablation experiments, and UMAP visualization. Specifically, welding upper surface molten pool images from both the source and target domains are fed into the backbone network, and the discrepancies between the predicted results and the true states are jointly evaluated using multiple loss functions. To the best of our knowledge, no UDA-based penetration prediction methods for different welding processes have been reported thus far. This study is applicable not only to classifying penetration state in welding but also to intelligent monitoring in other domains. Potential applications include additive manufacturing, defect detection, and computational materials science.

# 2. Experimental details

## 2.1 Experimental system

### 2.1.1 TIG welding experimental platform

The TIG welding experimental platform employed in this study comprises a TIG welding system, a motion control system, and a visual signal sensing system. The hardware configuration of the TIG welding system mainly consists of a welding power supply, a welding torch, and a wire feeder. The welding power supply utilized is the TETRIX-351 model from EWM. An STM32 microcontroller governs the movement of the welding torch, the platform's operation, and the wire feeding process. It independently regulates the welding speed, torch orientation, and wire feeding rate with high precision. During the welding process, the workpiece follows a predefined trajectory along the X-axis relative to the welding torch. Furthermore, the initiation of the welding arc and the transmission of motion control signals are achieved by a serial interface connecting the industrial computer to the STM32 controller.

The visual signal sensing system comprises a CMOS industrial camera (model: GS3-U3-41C6NIR), a 35 mm focal length lens, a compound filtering system, and an image acquisition module. The angle between the camera and the welding direction is adjustable to accommodate various operational requirements. To effectively reduce interference from welding plasma radiation, the system uses a compound filter assembly. This assembly consists of a 1064 nm narrowband filter (60 nm bandwidth, 0.01 % cutoff depth) cascaded with a neutral density filter (50 % transmittance). As illustrated in **Fig. 1**, the camera captures dynamic images of the molten pool in real time at a frame rate of 90 frames per second and transmits the captured data instantaneously to an industrial computer for storage and processing via the image acquisition system.

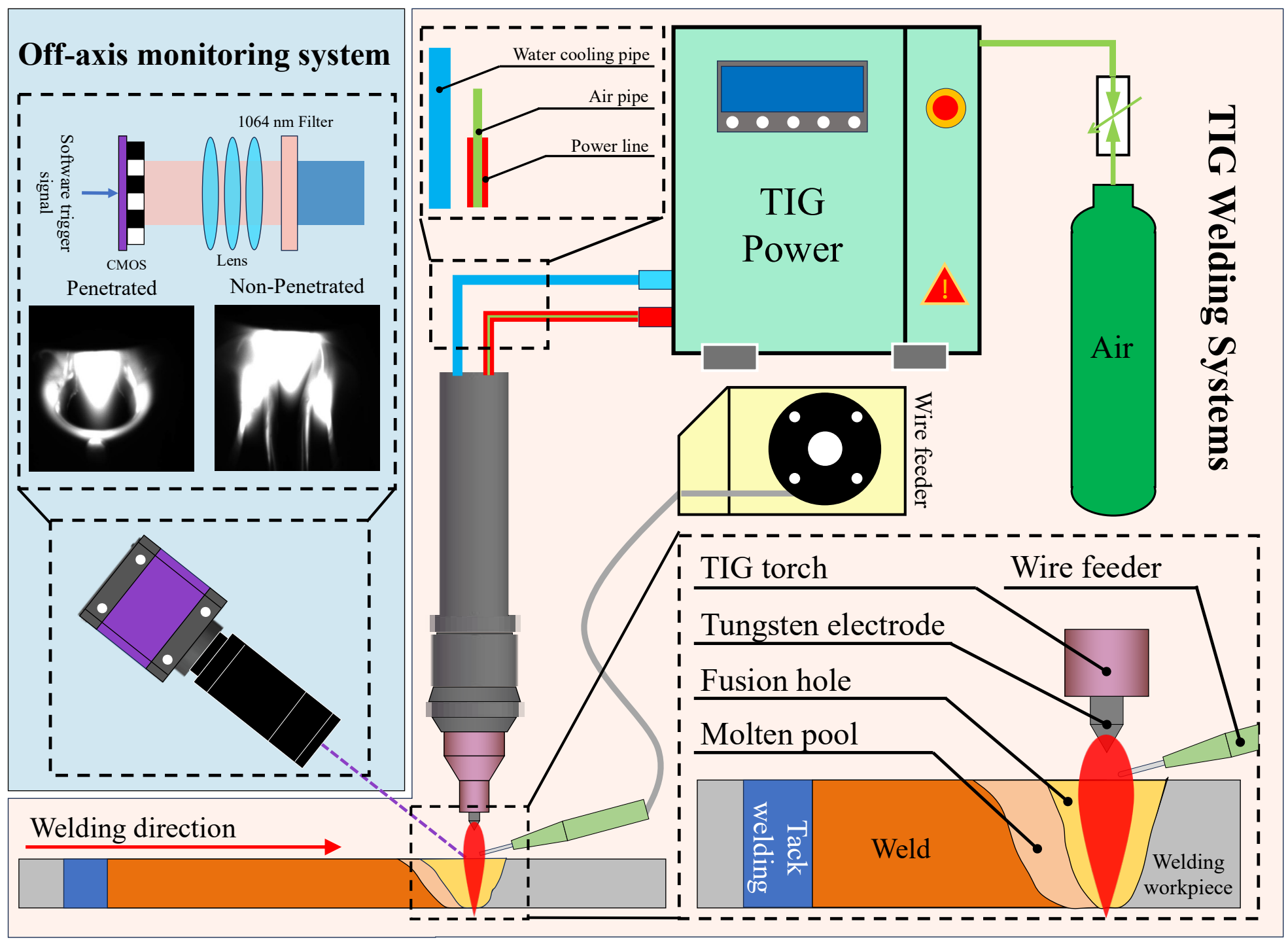

**Fig. 1** Schematic diagram of TIG welding inspection system.

2.1.1 laser welding experimental platform

The laser welding experimental platform is composed of three main modules: the laser welding system, the robotic motion control system, and the image acquisition system. The laser welding system includes a fiber laser, a water-cooling unit, and a laser head. The movement of the laser head is precisely controlled by a Yaskawa robot, which features a repeat positioning accuracy of 0.07 to 0.1 mm and enables laser welding at various locations through programming using a teach pendant. The output power of the fiber laser is regulated by the host computer, whereas other parameters, such as welding speed and defocus distance, are adjusted via the robotic control system. This study uses the YLS-20000 point-ring fiber laser. It has a maximum output power of 20 kW and operates at a 1080 nm wavelength. The laser is focused using a lens with a focal length of 165 mm. The beam parameter product (BPP) of the point laser is $3.385\,mm \cdot mrad$, resulting in a focused spot diameter of $197.535\,\mu m$ and a Rayleigh length of $2.88\,mm$. In contrast, the ring laser exhibits a BPP of $15.377\,mm \cdot mrad$, a focused spot diameter of $596.364\,\mu m$, and a Rayleigh length of $26.25\,mm$.

The laser experimental system integrates two CMOS cameras. The first camera (CMOS1) is configured with a coaxial optical path, oriented perpendicular to the welding workpiece. It enables real-time monitoring of the upper surface morphology and characteristics of the molten pool during the welding process. This camera is equipped with a flat light source emitting at a wavelength of 808 nm for auxiliary illumination. The second camera (CMOS2) adopts an off-axis optical path design, forming a 35° angle with the laser beam, and is employed for real-time observation of the upper surface morphology of the molten pool. This camera is paired with an auxiliary light source operating at a wavelength of 455 nm and incorporates a narrowband filter with a central wavelength of 455nm, a bandwidth of 60nm, and a cutoff depth of 0.01 %. Both CMOS cameras are set to an image acquisition frame rate of 200 FPS. To ensure the stability of the auxiliary light source intensity and enhance image quality, this study implements a PWM wave generator for synchronized control of the CMOS cameras and the auxiliary light sources. The PWM wave generator produces four identical PWM signals, two for driving the auxiliary light sources and the other two for triggering the CMOS cameras. A schematic diagram of the entire laser welding system is illustrated in Fig. 2.

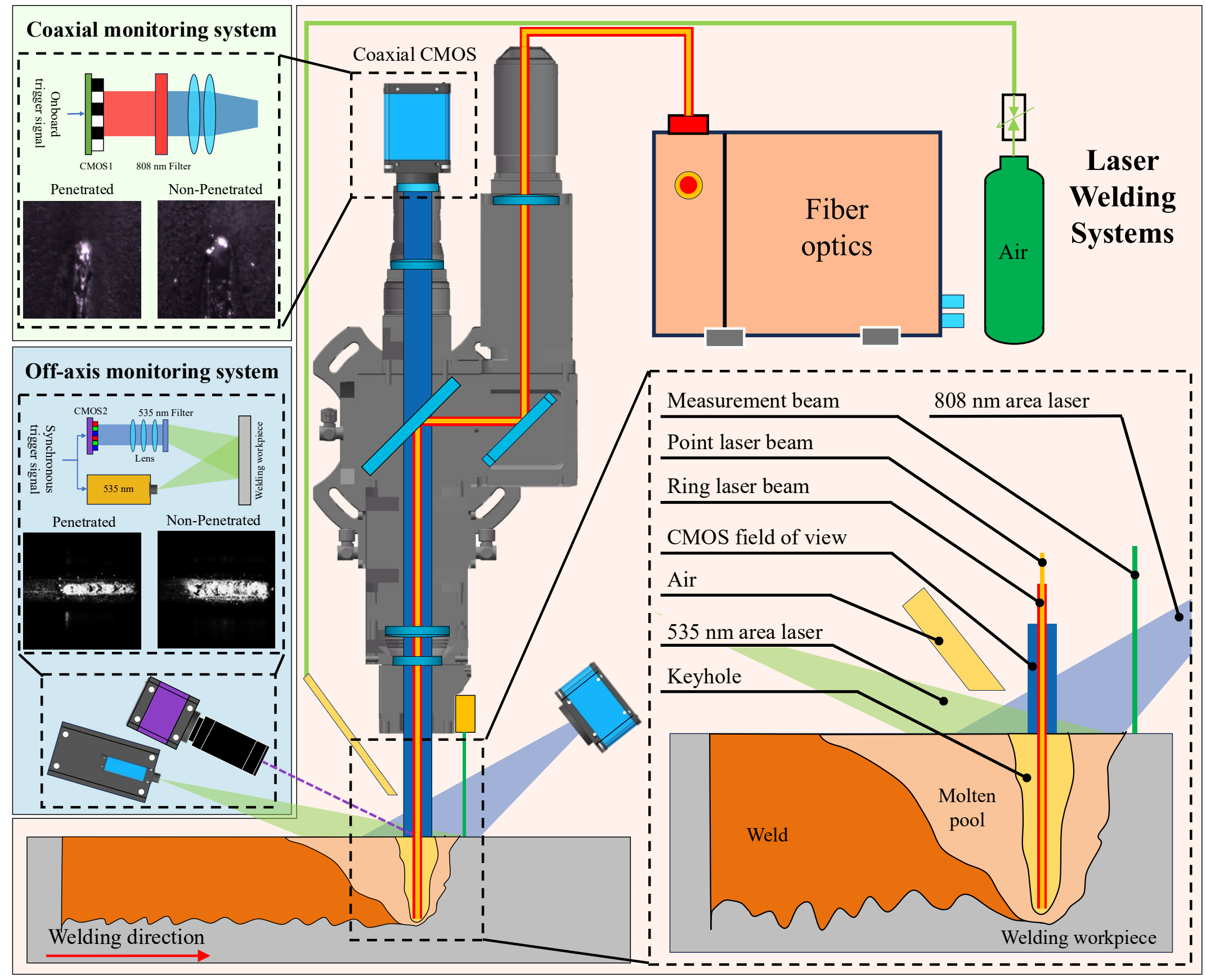

**Fig. 2** Schematic diagram of laser deep penetration welding inspection system.

## 2.2 Experimental method

### 2.2.1 TIG welding material

TIG welding was performed on SUS304 stainless steel specimens with dimensions of 100 mm × 40 mm × 3 mm (length × width × height). The welding configuration consisted of a butt joint with a predefined gap between the two workpieces, as depicted in **Fig. 3**. During the welding process, a cerium-doped tungsten electrode with a diameter of 3.2 mm and ER308L filler wire with a diameter of 1.2 mm were employed. High-purity argon gas (⩾99.99 %) was utilized as the shielding gas, with flow rates set to 15 $L/min$ for the front protection and 7.5 $L/min$ for the rear protection. The distance between the tungsten electrode and the workpiece was maintained at 3 mm. The chemical compositions of both the base material and the filler wire are detailed in **Tab. 1**.

**Tab. 1** The chemical composition of SUS304 and 308L welding wire (wt. %)[17]

| | C | Si | Mn | S | P | Cr | Ni | Fe |
|---|---|---|---|---|---|---|---|---|
| SUS304 | 0.06 | 0.53 | 1.00 | 0.003 | 0.035 | 18.23 | 8.19 | Bal. |
| ER308L | 0.019 | 0.46 | 1.72 | 0.003 | 0.013 | 20.80 | 10.10 | Bal. |

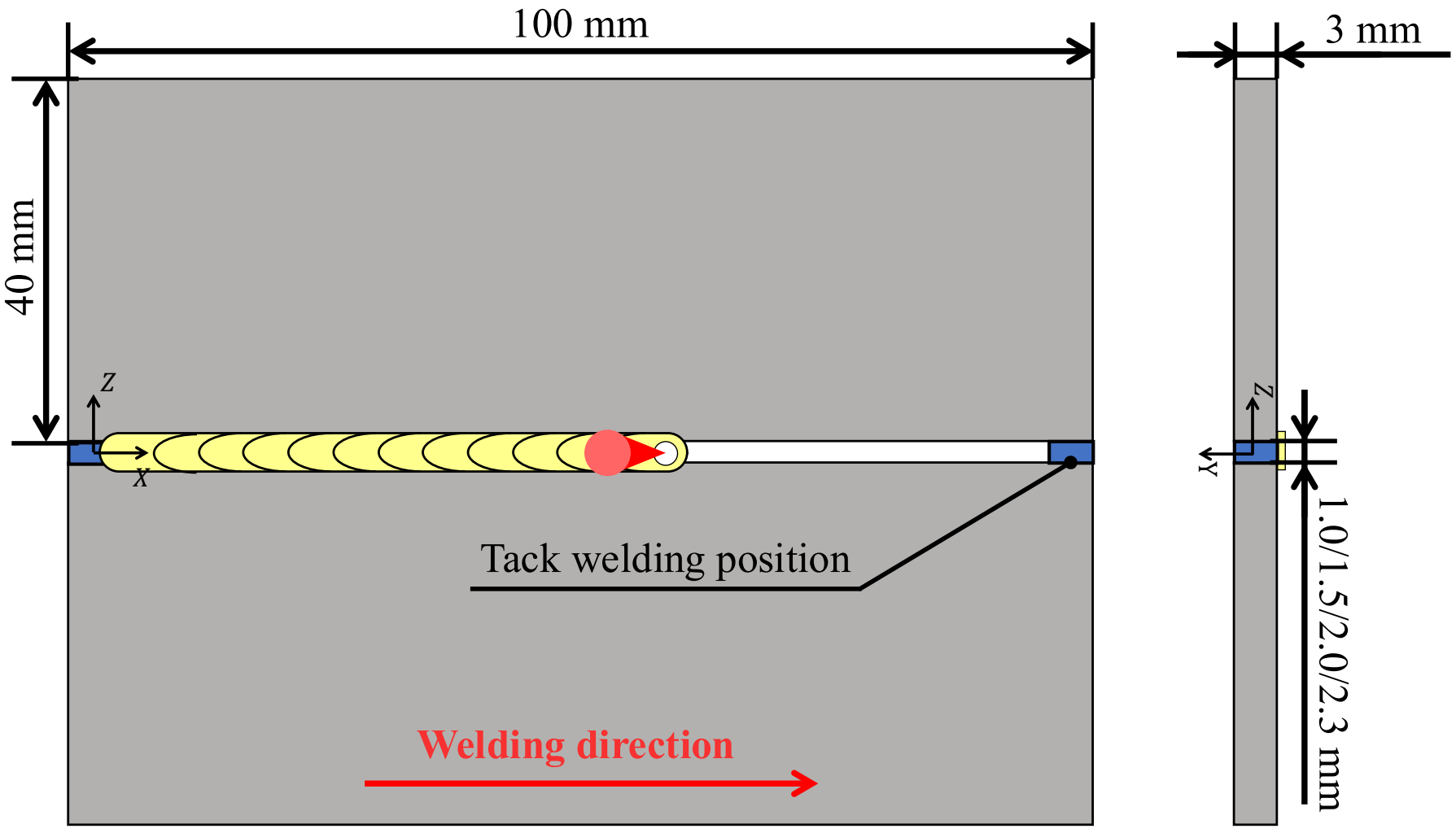

**Fig. 3** Dimensions of samples and grooves for the TIG welding experiment.

2.2.2 Laser welding material

The specimens utilized in the laser welding experiments were fabricated from COST-E material with dimensions of 100 mm × 40 mm × 50 mm (length × width × height). The geometric configuration of the workpieces consisted of a butt joint with a predefined V-groove and varied welding thicknesses, as depicted in **Fig. 4**, while their chemical composition is detailed in **Tab. 2**. During the welding process, the specimens remained stationary as the laser head moved along the weld seam direction. Shielding gas was delivered through three copper tubes arranged in parallel. The argon gas flow rate was set to 25 $L/min$ with a purity of 99.999 %. The lower edge of the protective gas nozzle was positioned 5 mm above the workpiece surface and 3 mm away from the laser beam, forming a 30° angle relative to the laser beam.

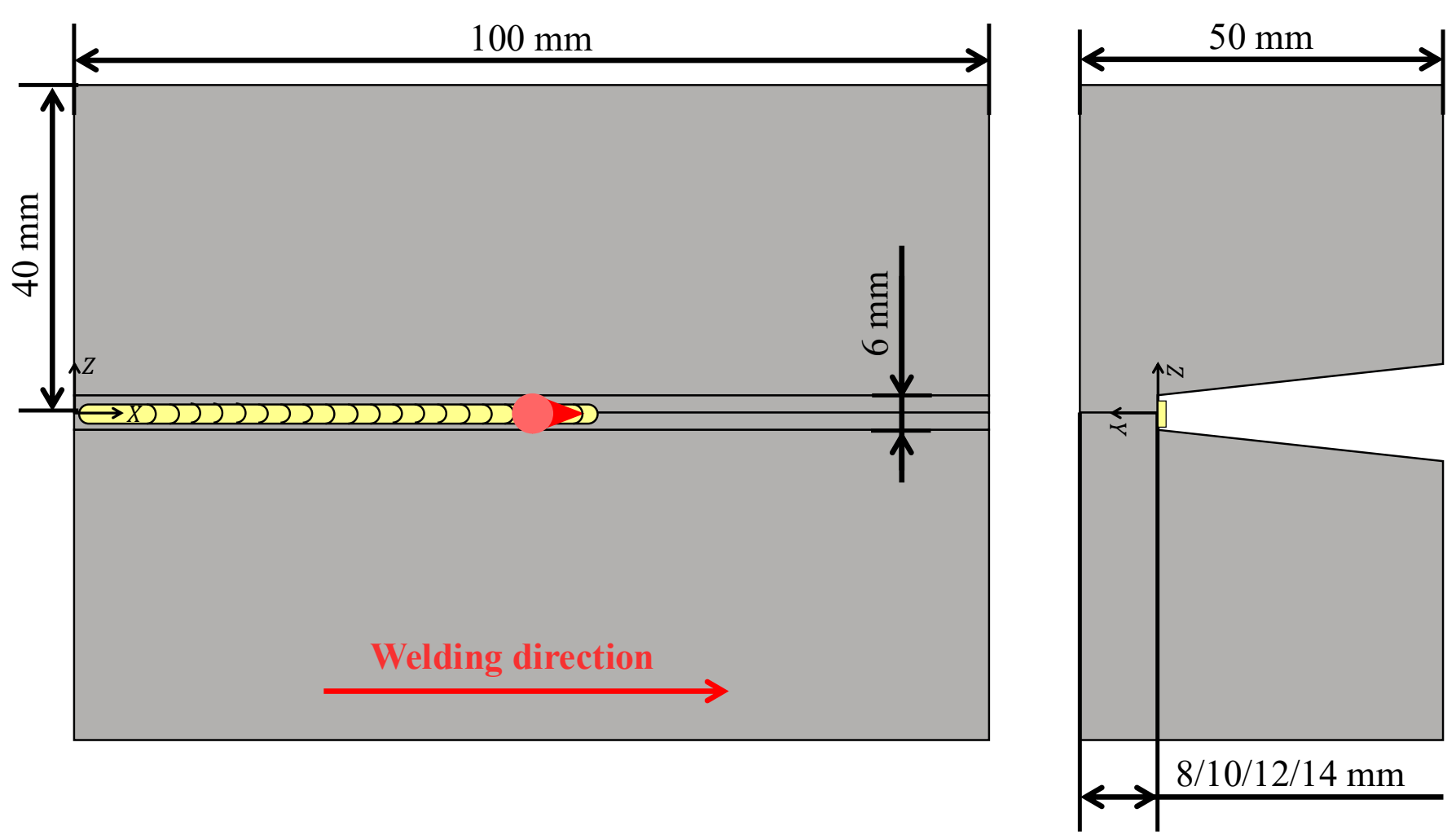

**Fig. 4** Dimensions of samples and grooves for the Laser welding experiment

**Tab. 2** Chemical composition of COST-E [18].

| COSTE | C | Mn | Si | Ni | Cr | Mo | W | Nb | V | Fe |
|---|---|---|---|---|---|---|---|---|---|---|
| (%) | 0.12 | 0.45 | 0.1 | 0.75 | 10.5 | 1.0 | 1.0 | 0.05 | 0.2 | Bal. |

## 2.3 Welding Datasets

Due to the geometric characteristics of the welded components, a through-hole forms during the TIG welding process. The presence of this hole can be used to determine whether reliable single-sided welding with double-sided formation has been achieved[19]. The molten metal, influenced by arc force and surface tension, surrounds the arc, thereby creating a fusion hole. In the images, this is represented as a low-intensity region located between a downward-pointing triangular high-intensity area (the arc) and an approximately circular high-intensity area (the molten metal). Furthermore, beneath the semi-circular molten pool, there exists a small rectangular high-intensity region, which results from the reflection of molten metal that solidifies later within the weld seam. Due to the angular position between the CMOS sensor and the welding gun, the upper edge of the fusion hole corresponds to the projection of the boundary line of the lower surface fusion hole of the workpiece onto the camera's imaging plane, while the lower boundary represents the projection of the boundary line of the upper surface fusion hole onto the imaging plane. When the fusion hole is absent, the molten metal accumulates only at the edges of the workpieces, leaving gaps unfilled and resulting in welding defects. Consequently, TIG welding can be classified into two primary states: the penetration state, characterized by a stable fusion hole, and the non-penetration state, marked by the disappearance of the fusion hole. These two states collectively constitute the classification dataset for TIG welding. The images captured during the TIG welding process are categorized into three sub-datasets, denoted as TIGFH-01, TIGFH-02, and TIGFH-03 in this study. The welding parameters include welding current ($I$), welding speed ($V_w$), workpiece thickness ($T$), workpiece gap ($Gap$), and wire feed rate ($WFR$). The welding parameters and camera settings for the three sub-datasets are presented in **Tab. 3**.

**Tab. 3** The experimental, equipment and parameter conditions of TIG.

| | TIGFH-01[20] | TIGFH-02[21] | TIGFH-03[22] |
|---|---|---|---|
| $I$ (A) | [85.0,90.0,95.0,100.0] | [130.0,135.0,140.0] | [120.0,130.0,140.0] |
| $V_w$ (m/min) | [0.07,0.08,0.09,0.10,0.11,0.12] | 0.20 | [0.18,0.20,0.22] |
| $T$ (mm) | 3 | 3 | 3 |
| $Gap$ (mm) | 1.0 | [1.5,2.0] | [1.0,1.5,2.0,2.3] |
| $WFR$ (m/min) | N/A | 0.75 | [0.5,0.6,0.7,0.8] |
| Shielding gas flow rate (L/min) | 10.0 | 10 | 15.0 |
| Camera angle (°) | 45.0 | 50.0 | 60.0 |
| Filter scheme | 1064 nm NBF FWHM: 30 nm | 1064 nm NBF FWHM: 30 nm | 1064 nm NBF FWHM: 30 nm |
| Exposure time (μs) | 5500 | 6500 | 8000 |
| Gamma | 1.5 | 1.45 | 1.53 |
| Black level (%) | 1.1 | 1.0 | 1.95 |
| Dataset samples | 4880 | 4880 | 4760 |

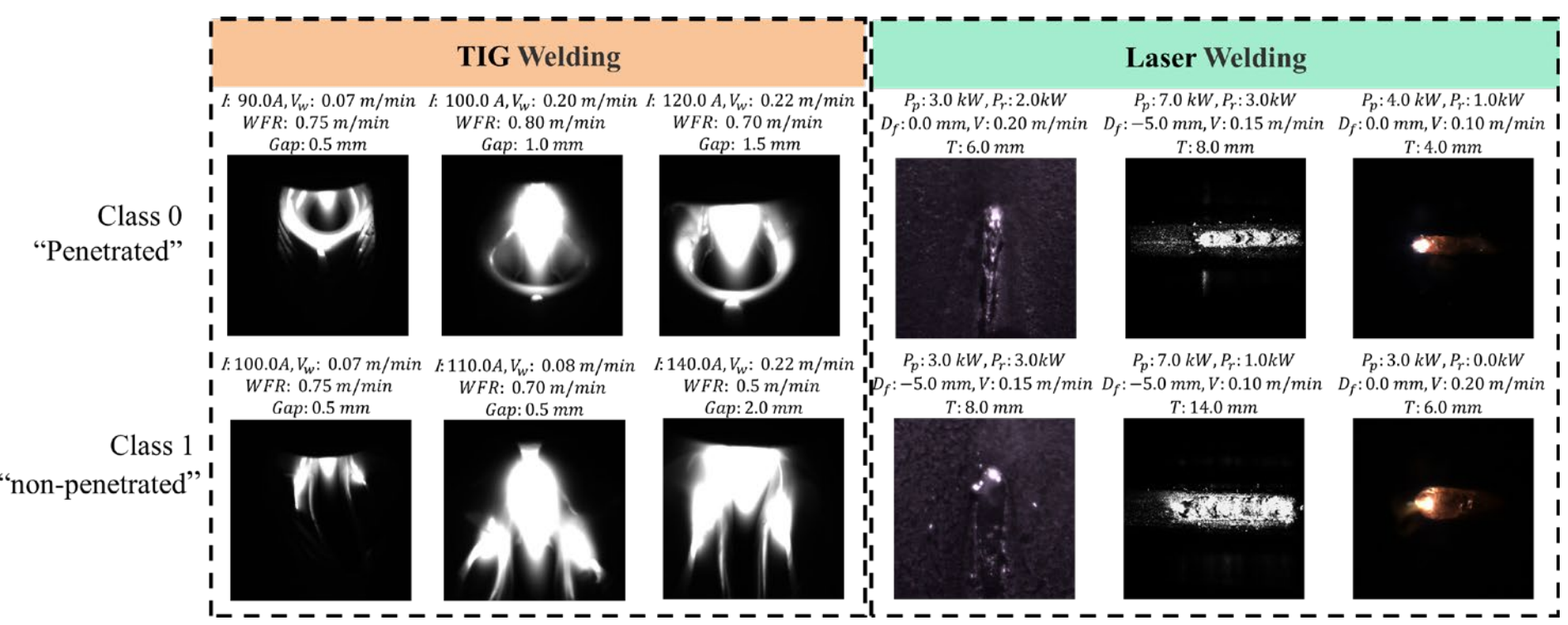

**Fig. 5** Images in the TIGFH and LSPS datasets.

There are three types of images acquired during laser welding, which are designated as LSPS-01, LSPS-02, and LSPS-03 in this study. The images in the LSPS-01 dataset were captured using CMOS1. The LSPS-02 images were obtained with CMOS2. While the LSPS-03 images were also captured by CMOS2 but under different conditions. Specifically, for LSPS-03, the images were acquired without an external light source or a narrowband filter, allowing the camera to directly record color images of the molten pool location. Welding parameters include spot laser power ($P_d$), ring laser power ($P_r$), defocusing amount ($D_f$), welding speed ($V_w$), and workpiece thickness ($T$). Detailed welding parameters and camera settings are presented in **Tab. 4**. Based on the aforementioned analysis, both TIG and laser welding processes are categorized into two primary states: penetration and non-penetration, corresponding to the presence or absence of a stable fusion hole/keyhole within the molten pool. Therefore, this study uniformly classifies the laser welding and TIG welding processes into two categories: penetrated and non-penetrated. **Fig. 5** presents images from the two datasets.

**Tab. 4** The experimental, equipment and parameter conditions of Laser weld.

| | LSPS-01 | LSPS-02 | LSPS-03 |
|---|---|---|---|
| $P_d$ (kW) | [3.0,4.0,5.0,6.0] | [7.0,8.0,9.0] | [3.0,4.0,5.0] |
| $P_r$ (kW) | [0.0,1.0,2.0,3.0,4.0] | [0.0, 1.0, 2.0, 3.0] | [0.0, 1.0] |
| $D_f$ (mm) | [0.0, -5.0] | [5.0, 0.0, -5.0, -7.0] | 0.0 |
| $V_w$ (min/min) | [0.10,0.15,0.20] | [0.10,0.15,0.20] | [0.10,0.20] |
| $T$ (mm) | [4.0,6.0,8.0,9.0,10.0] | [8.0,10.0,12.0,14.0] | [4.0,6.0] |
| Camera sensor | 1/4″ Mono CMOS | 1/2.9" COLOR CMOS | 1/2.9" COLOR CMOS |
| Camera angle (°) | 0.0 | 0.0 | 60.0 |
| Filter scheme | 808 nm NBF<br>FWHM: 30 nm | 455 nm NBF<br>FWHM: 30 nm | N/A |
| Exposure time (μs) | 5 | 5 | 36 |
| Gamma | 1.0 | 1.5 | 1.5 |
| Black level (%) | 1.0 | 1.95 | 1.95 |
| Dataset samples | 4530 | 4782 | 4342 |

# 3. Model architecture

The overall architecture of the UDA training network is illustrated in **Fig. 6**.

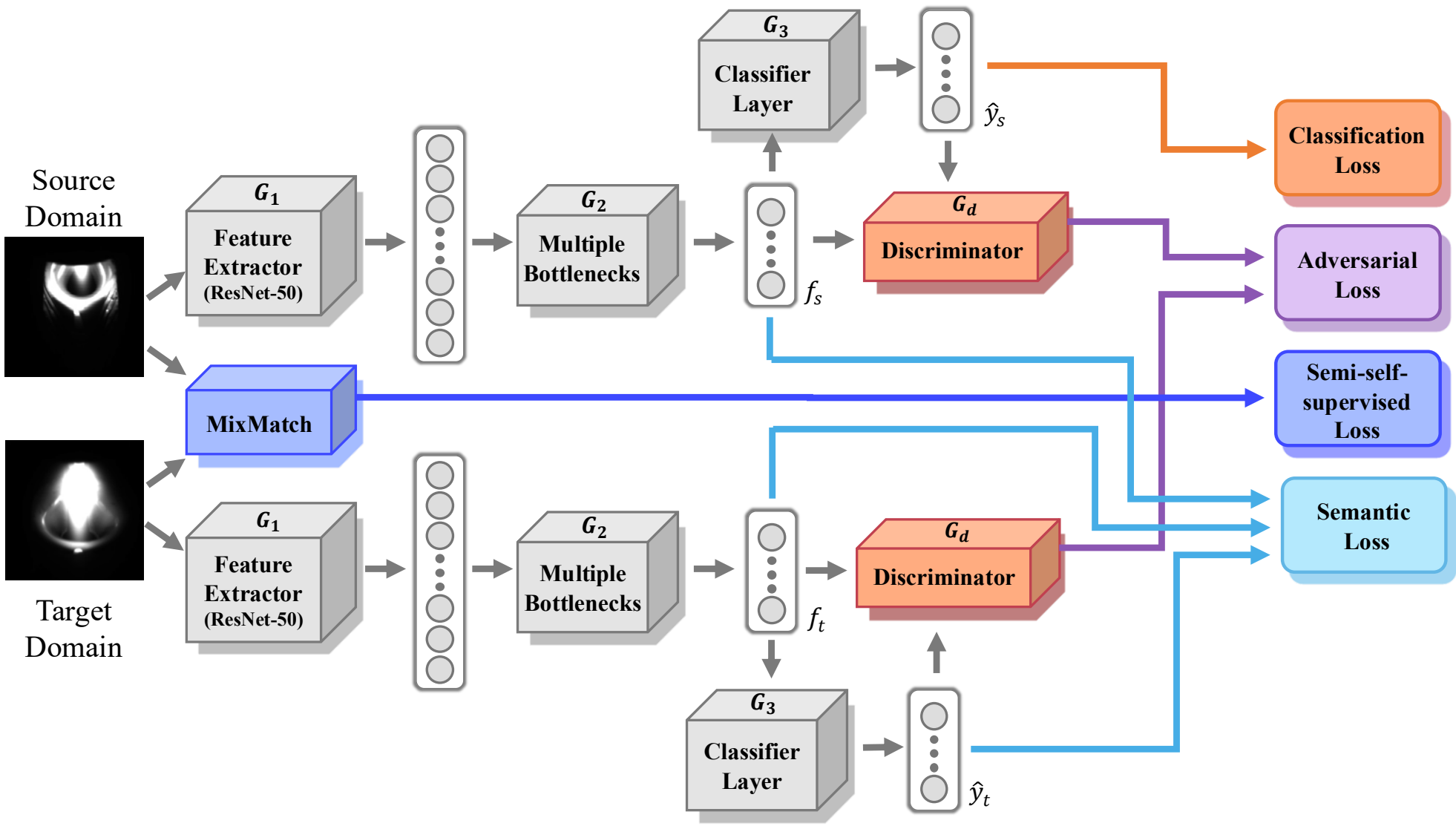

**Fig. 6** The overall structure of the UDA model with three new loss functions

### 3.1 Unsupervised domain adaptation

In unsupervised domain adaptation, the primary objective is to minimize the discrepancy between the source domain and the target domain. For the source domain $D_s$, a set of labeled samples $D_s = (x_{I,s}, y_{I,s})_{i=1}^{n_s}$ is provided, where $x_{I,s}$ denotes a sample and $y_{I,s}$ represents its corresponding label. Conversely, the target domain $D_t$ consists only of unlabeled samples $D_t = (x_{I,t})_{i=1}^{n_t}$. The goal is to estimate the target domain labels $\hat{y}_{I,t}$ by using a shared feature space. This process must account for both the similarities and differences between the source and target feature spaces. This study focuses on a closed-set setting, where the source domain and target domain share identical label categories, denoted as $C_s = C_t$. When applied to laser welding images and TIG welding images, this corresponds to classifying lack of penetration versus penetration in both the source and target domain images. Notably, in UDA methods, the labels of target domain data are utilized exclusively for evaluating model performance and are not required during the training process, thereby facilitating broader deployment and application of the model. Additionally, the molten pool feature transfer method proposed in this paper can also be extended to scenarios involving labeled molten pool images, specifically in SDA and SSDA.

This study employs the GSDE strategy, which involves iteratively training the model multiple times. In each iteration, the network training is restarted from scratch with reinitialized weights. Additionally, the number of pseudo-source samples is progressively increased to incorporate prior knowledge acquired in the previous iteration. Specifically, the strategy trains the network for $N$ iterations, where in each iteration, the network weights are reinitialized, and the source domain is expanded by incorporating target domain samples: $D'(n)_s = D(s) \cup D(n)'_t$, where $D(n)'_t$ is a subset of $D(t)$. Based on the prediction results from the preceding iteration, pseudo-labels are assigned to the samples in the subset $D(n)'_t$, which can be expressed as: $D(n)'_t = (x_{I,t}, y_{I,t})_{i=1}^{n_t}$.

This training strategy allows the use of substantial target domain data as pseudo-source data. However, it introduces a potential drawback: the risk of adding misclassified samples to the pseudo-source dataset. Such errors are undoubtedly detrimental to domain adaptation. To mitigate this issue, the present study selectively incorporates samples with the highest prediction confidence as pseudo-source data, thereby reducing the likelihood of using incorrectly classified samples. Specifically, during the $n$-th training iteration, the top $(n-1)/N$ proportion of data samples with the highest confidence are selected, and their pseudo-labels are added to the source domain as pseudo-source data. **Alg. 1** presents the pseudo-code of the GSDE algorithm.

**Alg. 1** Algorithm of the proposed gradual source domain expansion.

**01:** Initialize pseudo-labeled target set $D_t' = \emptyset, D_s' = D_s$
**02: for** n=1 **to** N **do** // *GSDE Iteration Loop*
**03:** Initialize weights of $G_1, G_3, G_3, G_d$
**04: for** e=1 **to** E **do** // *Epoch Training Loop*
**05: for** each batch $\{(x_s, y_t), x_t\}$ from $D_s'$ and $D_t$ **do**
**06:** // *Feature Extraction*
**07:** $f_s = G_2(G_1(x_s)), f_t = G_2(G_1(x_t))$
**08:** $p_s = G_3(f_s), p_t = G_3(f_t)$
**09:** // *Loss Calculation*
**10:** $L = L_{CLAS}(p_s, y_s) + L_{ADV}(f_s, p_s, f_t, p_t, G_d) + L_{MS}(f_s, p_s, f_t, p_t) + L_{SS}(x_s, y_t, x_t)$
**11:** Update weights of $G_1, G_3, G_3, G_d$ by backpropagating $L$
**12: end for**
**13: end for**
**14: if** $n < N$ **then**
**15:** $AllScores = []$
**16: for** each $x_{ti} \in D_t$ **do**
**17:** $p_{ti} = G_3(G_2(G_1(x_{ti})))$
**18:** $score_i = ConfidenceScore(p_{ti})$ // using probability score and Eq.7
**19:** $AllScores.append(score_i)$
**20: end for**
**21:** $D_{inter}' = D_t \in Top\frac{n-1}{N} of\ AllScores, D_t' = \emptyset$
**22: for** each $x_i \in D_{inter}'$ **do**
**23:** $\hat{y}_i = arg\ max(G_3(G_2(G_1(x_i))))$
**24:** Add $(x_i, \hat{y}_i)$ to $D_t'$
**25: end for**
**26: end if**
**27: end for**
**28: Return** $G_1, G_2, G_3$

In domain adaptation tasks, two prevalent strategies are adversarial adaptation [23] and entropy minimization adaptation [24]. Adversarial adaptation aims to generate domain-invariant features through adversarial training, thereby aligning the feature spaces of the source and target domains. Entropy minimization adaptation, conversely, focuses on reducing the predictive entropy of target samples by increasing the model's predictive confidence for target domain data. This drives sample points in the target domain away from the classification decision boundary, consequently improving the model's domain adaptation performance. The fundamental principles of GSDE are illustrated in **Fig. 7** (Adversarial Adaptation) and **Fig. 8** (Entropy Minimization). Initially, the feature distributions of the source and target domains exhibit significant domain shift. Through the adversarial training mechanism, the feature extractor progressively learns to generate domain-invariant feature representations. Mathematically, this process can be interpreted as the progressive optimization of statistical distance metrics between the feature spaces of the source and target domains, ultimately achieving cross-domain feature distribution alignment. As depicted in the figures, the samples of the source and target domains gradually converge within the same feature space. It is crucial to emphasize that adversarial adaptation essentially represents a class-agnostic global alignment strategy. As shown in **Fig. 7(b),** despite the feature spaces of the two domains exhibiting a tendency toward consistency in their overall distribution, the absence of fine-grained class constraints may result in some target domain samples being misclassified due to falling within the misjudgment region of the source domain classification boundary. This indicates that achieving mere alignment of feature space does not ensure the consistency of the conditional distributions $P(y|x_s)$ and $P(y|x_t)$. Consequently, a pseudo-source data guidance mechanism is introduced. The core implementation workflow is shown in **Fig. 7(c)** and **(d)**. By creating a dynamic pseudo-source domain using high-confidence pseudo-labeled samples and their corresponding pseudo-labels screened during the preceding iterative training process, class-conditional anchor points are established within the feature space

The main challenge in entropy minimization adaptation involves the misclassification of the decision boundary within the target domain, as illustrated in **Fig. 8(a)**. This issue frequently leads to the accumulation of early-stage errors, especially for samples located near the decision boundary (**Fig. 8(b)**). By introducing high-confidence pseudo-source labels, the alignment of the decision boundary can be effectively guided across both the source and target domains, thereby ensuring classification consistency (**Fig. 8(c)**). As a result, the model demonstrates superior performance during the initial stages of training (**Fig. 8(d)**). In summary, GSDE combines adversarial adaptation with entropy minimization strategies, progressively incorporating high-confidence pseudo-label samples to create a continuous optimization space that transitions from global alignment to fine-grained category alignment. Additionally, pseudo-source samples mitigate error accumulation and enhance the model's resilience to noise in the target domain.

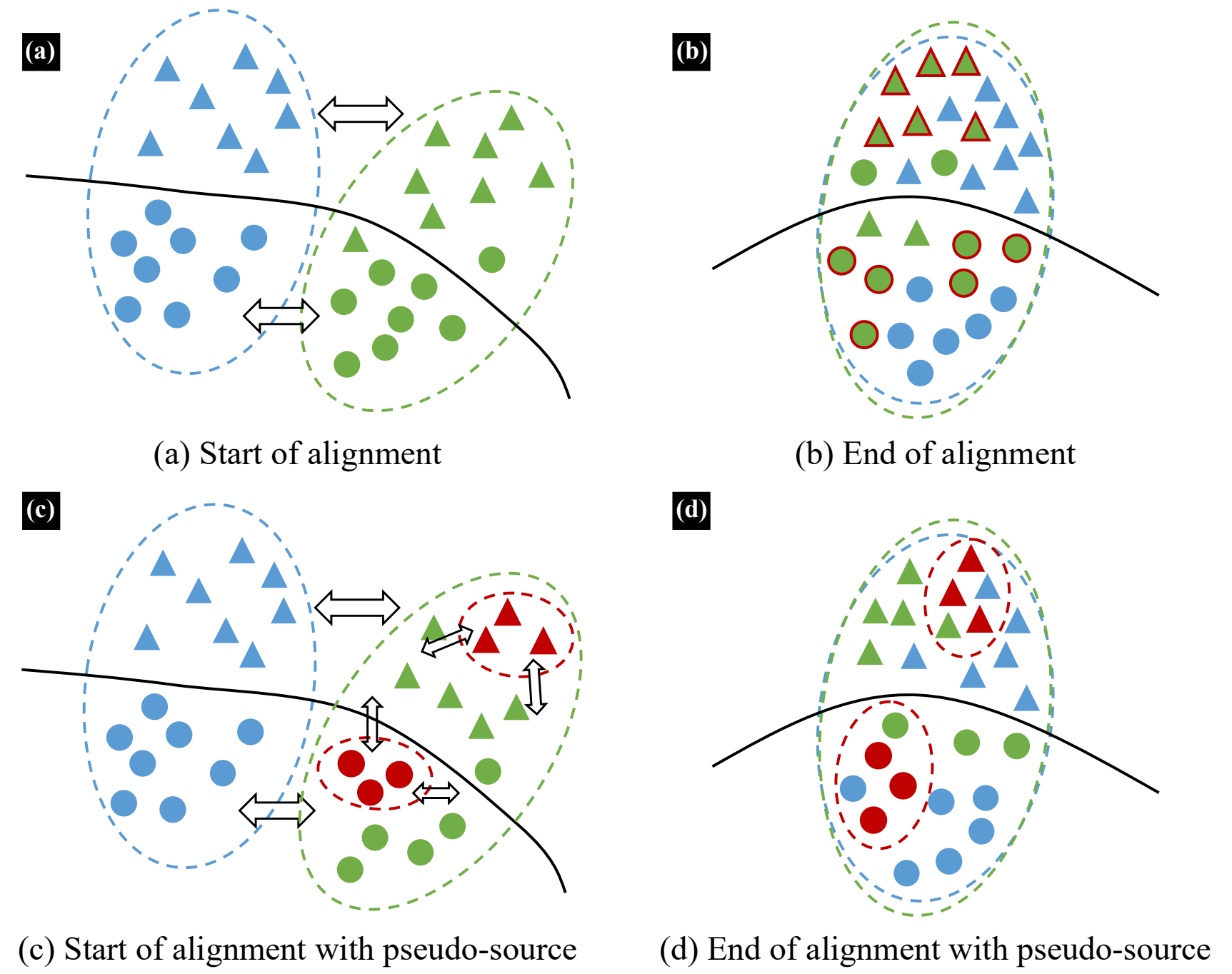

**Fig. 7** Schematic of adversarial adaptation. During the alignment of the source and target domains (a), some samples may be misclassified (b). In a new round of training, using high-confidence target domain samples as pseudo-source samples (c) helps to guide the adaptation process, resulting in improved domain adaptation performance (d).

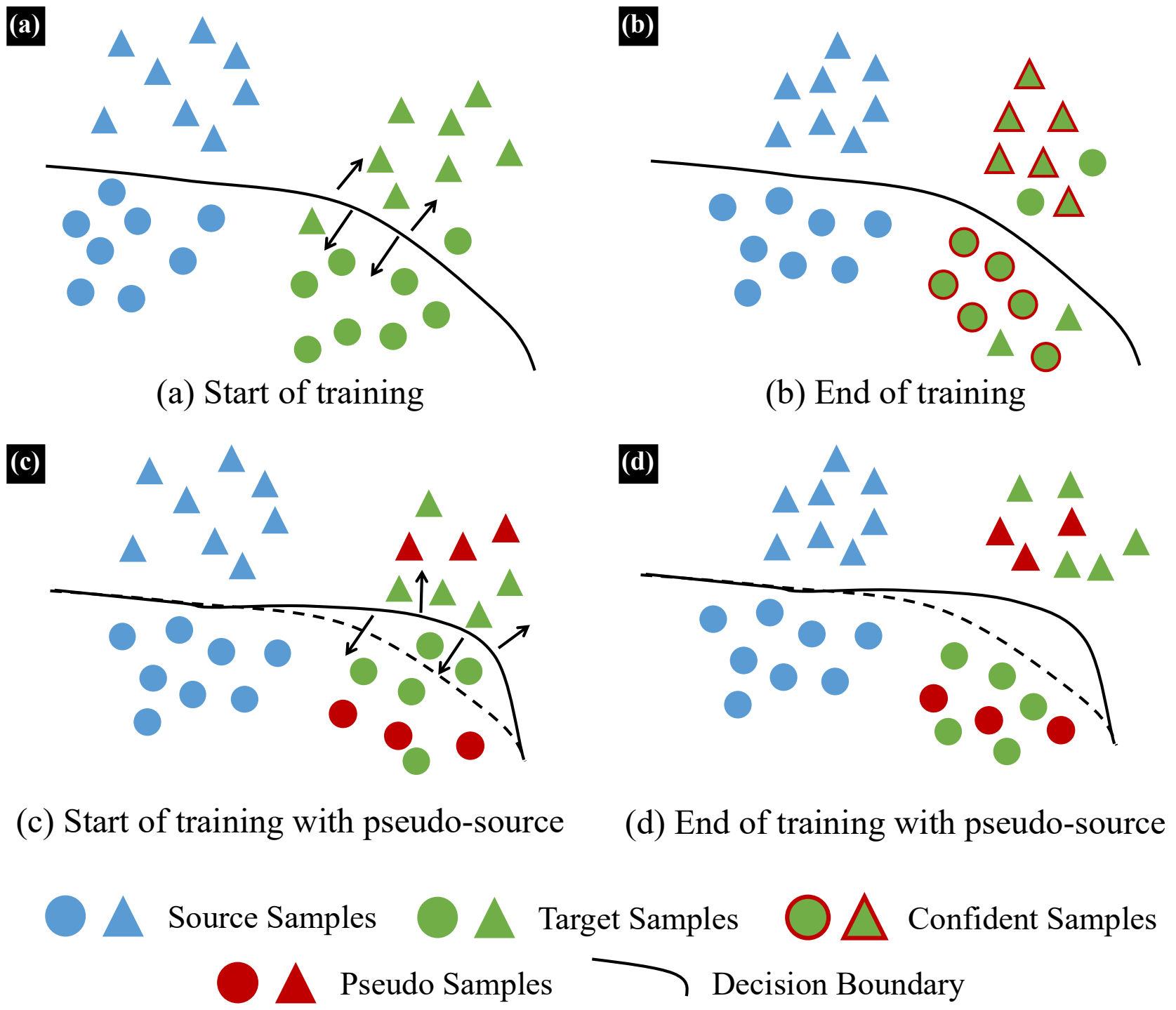

**Fig. 8** Schematic of entropy minimization adaptation. Because the classifier and target domain samples are not aligned (a), samples may ultimately be misclassified (b). Using high-confidence target domain samples as pseudo-source samples helps to guide the classifier (and its decision boundary) to adjust and adapt to the target domain samples (c), resulting in improved domain adaptation performance (d).

### 3.2 Transfer loss functions

This section elaborates on the four loss functions utilized in the transfer learning network introduced in this study, namely the classification loss $L_{CLAS}$, the adversarial loss $L_{ADV}$, the semantic loss $L_{MS}$, and the semi-supervised loss $L_{SS}$:

$$L = L_{CLAS} + L_{ADV} + L_{MS} + L_{SS} \tag{1}$$

Among these, the classification loss $L_{CLAS}$ is concurrently applied to both the source domain $D$ and the extended source domain $D'$. To enhance the model's transferability, it is a common approach to minimize the classification loss while incorporating additional loss functions. In this paper, the classification loss $L_{CLAS}$ is expressed as follows:

$$L_{CLAS} = \frac{1}{N_s}\sum_{i=1}^{N_s} L_{ce}\left(G_3\left(f_{i,s}\right), y_{i,s}\right) \tag{2}$$

where $L_{ce}$ denotes the cross-entropy loss, and $G_3$ represents the domain classification network.

Furthermore, this study incorporates three additional adaptive loss functions to provide supplementary support. The adversarial loss $L_{ADV}$ implicitly aligns the global distributions of the source and target domains through adversarial training. The semantic loss function $L_{MS}$ concurrently models both global distributions and local semantic relationships, such as inter-class similarities, thereby mitigating biases inherent in single alignment approaches. The semi-supervised loss functions $L_{SS}$ is utilized to enhance the model’s ability to extract invariant features from augmented images.

#### 3.2.1 Adversarial loss: $L_{ADV}$

The core concept of adversarial loss is to generate domain-invariant feature representations through adversarial training between the feature extractor ($G_1$, $G_2$) and a domain discriminator ($G_d$). This process involves two competing objectives. First, $G_d$ is trained specifically to distinguish the origin of feature representations (i.e., whether they are from the source or target domain). Concurrently, the feature extractor is trained to produce features that are indistinguishable to $G_d$. This is achieved by a gradient reversal layer, which inverts the backpropagated gradients from the discriminator. This mechanism compels the feature extractor to learn in a direction that maximizes $G_d$'s error, thereby forcing the alignment of feature distributions from both domains. Given the substantial discrepancies between the laser and TIG welding datasets, this adversarial loss was essential. In this study, the CDAN [25] network is utilized to construct $L_{ADV}$:

$$L_{ADV} = \lambda \cdot L_{BCE}\left(G_d\left(\left(f_i \otimes G_3(f_i)\right), d_i\right)\right) \tag{3}$$

where $f_i$ denotes the features of sample $i$ and $d_i$ is the domain label. Consistent with CDAN, this study also employs a gradually increasing scaling factor $\lambda$ to control the adversarial loss. A gradient reversal layer is inserted before the domain classification network, aiming to transform the training objective from distinguishing different domains into generating domain-invariant features that are challenging to discriminate.

### 3.2.2 Semantic loss: $L_{MS}$

In this study, the second loss function is the moving semantic transfer loss which was introduced in MSTN [26]. Within the feature space, each class is associated with a centroid that represents the mean feature vector of all samples belonging to that class. The semantic loss works by minimizing the distance between the centroids of corresponding classes in the source and target domains. This process ensures that these centroids remain close in the feature space. This methodology ensures that identical classes across different domains exhibit similar feature representations.

$$L_{MS} = \sum_{k=1}^{K} D\left(C_s^k, C_t^k\right) \tag{4}$$

where $C_s^k$ and $C_t^k$ represent the centroids of the *k*-th class in the source and target domains, respectively, and D is the squared Euclidean distance, defined as $D(x, x') = \|x - x'\|^2$. This approach explicitly minimizes the distance between the centroids of the same class to achieve class alignment. Inspired by recent deep clustering-based methods [27,28], Thomas et al. [29] further extended this loss term by also increasing the distance between centroids of different classes:

$$L_{MS} = \sum_{k=1}^{K} \lambda \cdot \Theta\left(C_s^k, C_t^k\right) + \sum_{k=1}^{K} \sum_{j \neq k}^{K} \Theta\left(C_s^k, C_s^j\right) + \lambda \cdot \Theta\left(C_s^k, C_t^j\right) + \lambda \cdot \Theta\left(C_t^k, C_t^j\right) \tag{5}$$

where $\Theta$ represents the cosine similarity function, which calculates the cosine of the angle between two vectors. A higher value signifies greater similarity between the vectors, whereas a lower value indicates lesser similarity. It is formally defined as: $(x, x') = \frac{x \cdot x'}{\|x\| * \|x'\|}$. The three newly introduced terms are:

**Intra-source-domain inter-class separation term $\sum\sum\Theta(C_s^k, C_s^j)$:** This term aims to maximize the distance between the centroids of different classes in the source domain, thereby enhancing the discriminative power of the source features.

**Cross-domain inter-class separation term $\Theta(C_s^k, C_t^j)$:** This term ensures that the centroids of different class in the source domain do not approach those in the target domain, thus avoiding potential cross-domain mismatches.

**Intra-target-domain inter-class separation term $\Theta(C_t^k, C_t^j)$:** This term implicitly encourages feature separation among different classes within the target domain.

### 3.2.3 Semi-self-supervised loss: $L_{SS}$

This study employs the MixMatch [30] as the loss function for semi-supervised learning. The algorithm integrates consistency regularization, MixUp regularization, and an entropy minimization strategy. Specifically, consistency regularization enhances model robustness by applying multiple data augmentations (such as flipping and cropping) to unlabeled data and constraining the model to produce consistent predictive distributions across different augmented versions of the same data. This constraint is enforced using an L2 loss function. MixUp regularization generates new training samples through linear interpolation between labeled and unlabeled data, thereby smoothing the decision boundaries and improving the model's generalization capabilities. Additionally, the entropy minimization strategy employs a sharpening operation to scale the predictive distributions of unlabeled data, reducing their entropy and guiding the decision boundaries away from high-density data regions.

### 3.3 Additional improvements

3.3.1 Multiple bottlenecks

Inspired by [31] , this study adopts a parallel multiple bottlenecks architecture. In the network structure, $k$ bottleneck structures are deployed in parallel, and a differentiated initialization strategy is utilized to perform a weighted average fusion of the outputs of each bottleneck. This multiple bottleneck collaborative mechanism improves model performance in two dimensions: first, each bottleneck within the parallel structure employs independently initialized parameters, thereby effectively breaking the symmetry constraints of the parameter space; second, the multi-path information processing mechanism promotes the model to explore a broader solution space, thus preventing the optimization process from becoming trapped in local minima. The input to each bottleneck is the feature map from the final pooling layer of the ResNet-50 backbone, with a shape of ($B \times C \times W \times H$). Each bottleneck processes this input into a feature vector of shape ($B \times C \times 1 \times 1$). The outputs from all $k$ bottlenecks are then concatenated and fused through a weighted average, yielding a final output of shape ($B \times C \times k$). The structure of the multiple bottlenecks is shown in **Fig. 9**. Mathematically, the architecture can be formally defined as follows:

$$G_2(y_b) = \frac{1}{k}\sum_{m} B_k(y_b) \tag{6}$$

where, $y_b$ represents the output of the backbone network, and $B_k$ denotes the $k$-th bottleneck.

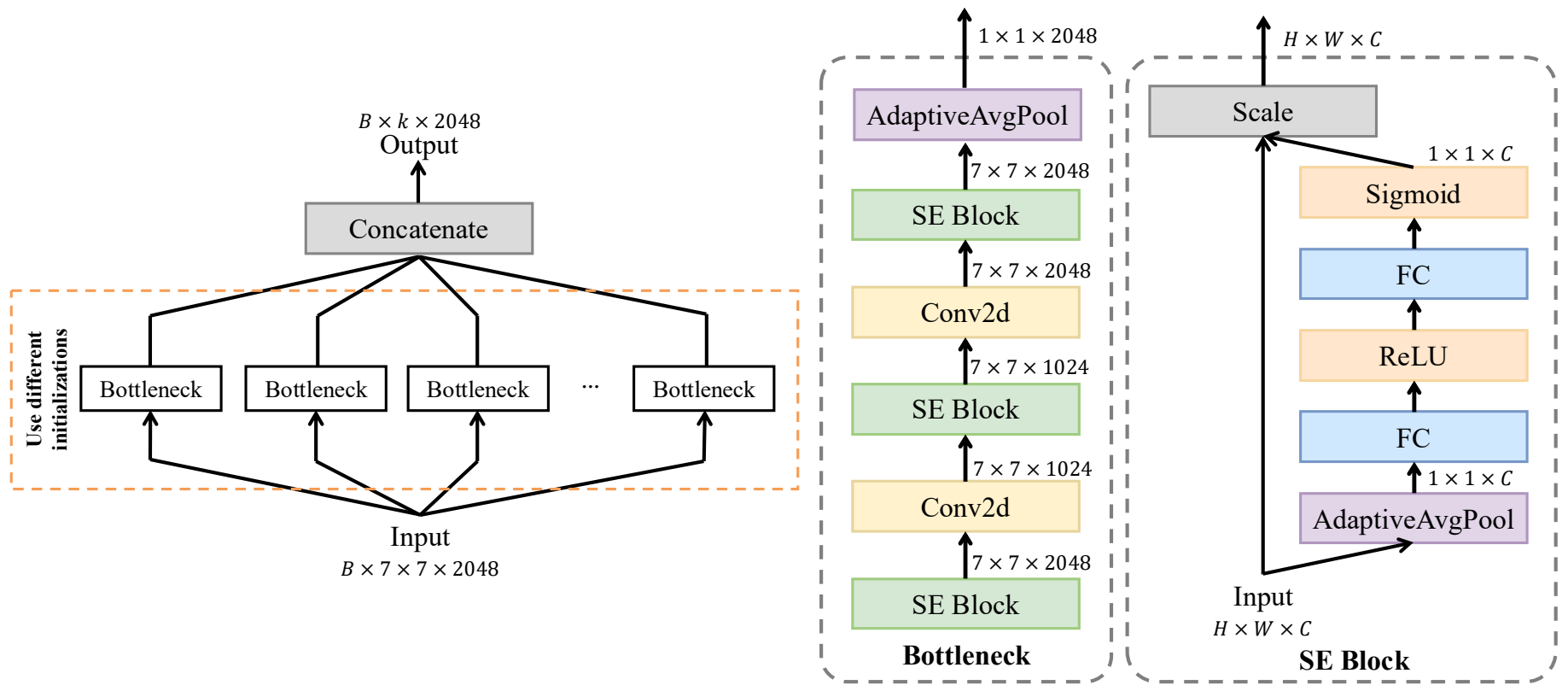

**Fig. 9** Schematics of the multiple bottlenecks

3.3.2 Scoring of pseudo labels

In this study, we also introduce the pseudo-label scoring algorithm presented in [29]: In addition to utilizing the classifier's probability score, this study further employs a neighborhood aggregation score:

$$p_i^{all} = \frac{1}{2}\left(p_i + p_i^{NA}\right) \tag{7}$$

The neighborhood aggregation score is obtained by identifying the *m* nearest neighbors in the feature space and aggregating their classification probability scores. Unlike methods that rely solely on class centroids, the neighborhood aggregation approach can more effectively capture local data structures, thereby improving the quality of the pseudo-labels. It can be expressed as:

$$p_i^{NA} = \frac{1}{m}\sum_{j\in N} p(j) \tag{8}$$

where $N$ is the set of all target data, and $p(j)$ represents the soft prediction of the $j$-th target data point, i.e., the probability distribution corresponding to that target data point.

# 4. Results and discussion

This study evaluates the proposed method using two welding process datasets. The experimental model is implemented within the PyTorch framework, with the training environment configured as follows: an Intel i7-13900KF CPU, a NVIDIA RTX 2080 Ti GPU, and 32GB of memory. The model's evaluation metric is accuracy ($Acc$), which is calculated as follows:

$$Acc = \frac{TP + TN}{TP + TN + FP + FN} \tag{9}$$

where TP (true positive), FP (false positive), FN (false negative), and TN (true negative) represent the number of correctly and incorrectly classified samples.

### 4.1 Baseline network

To systematically evaluate the performance of various transfer learning strategies, this study designed three baseline networks. Baseline-1 was trained exclusively on source domain data and subsequently evaluated on target domain data. Baseline-2 underwent initial pre-training on source domain data, followed by comprehensive fine-tuning of all parameters using target domain data. The final model performance was assessed jointly on both source and target domain datasets. Baseline-3 followed a similar training and evaluation procedure as Baseline-2; however, during the fine-tuning phase, only the parameters of the classification layer and the last convolutional layer were unfrozen, while the remaining network parameters remained frozen. Under identical welding conditions, the performances of the three baseline networks are summarized in **Fig. 10** and **Fig. 11**, with the changes in accuracy (presented in parentheses) indicating the relative improvements or declines compared to Baseline-1. The training parameters are delineated as follows: the dataset is partitioned into training and validation subsets in an 8:2 ratio; the optimizer employed is Adam [32] with an initial learning rate of 0.001 and the OneCycleLR strategy for dynamic adjustment; the batch size is established at 64; the number of training epochs is set to 200; and AMP training is activated to enhance computational efficiency, concurrently recording the best performance results on the validation set„All three baseline networks are trained using ResNet50 [33].

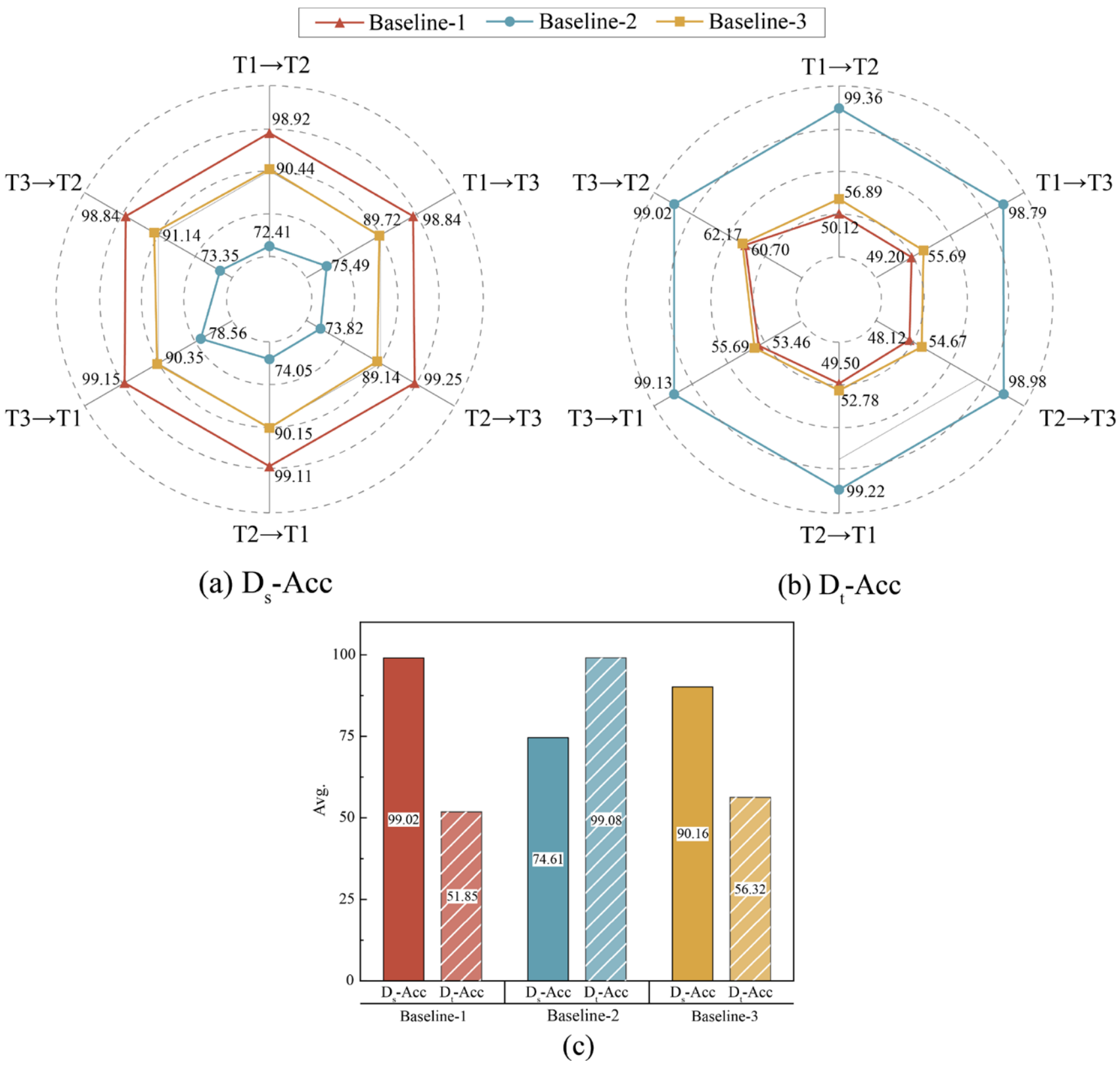

**Fig. 10** The baseline's accuracy results on TIGFH dataset. (a) The accuracy results on $D_s$. (b) The accuracy results on $D_t$.(c) The average accuracy results on TIGFH dataset. (For an explanation of the color references in the legend of this figure, please refer to the online version of this paper.)

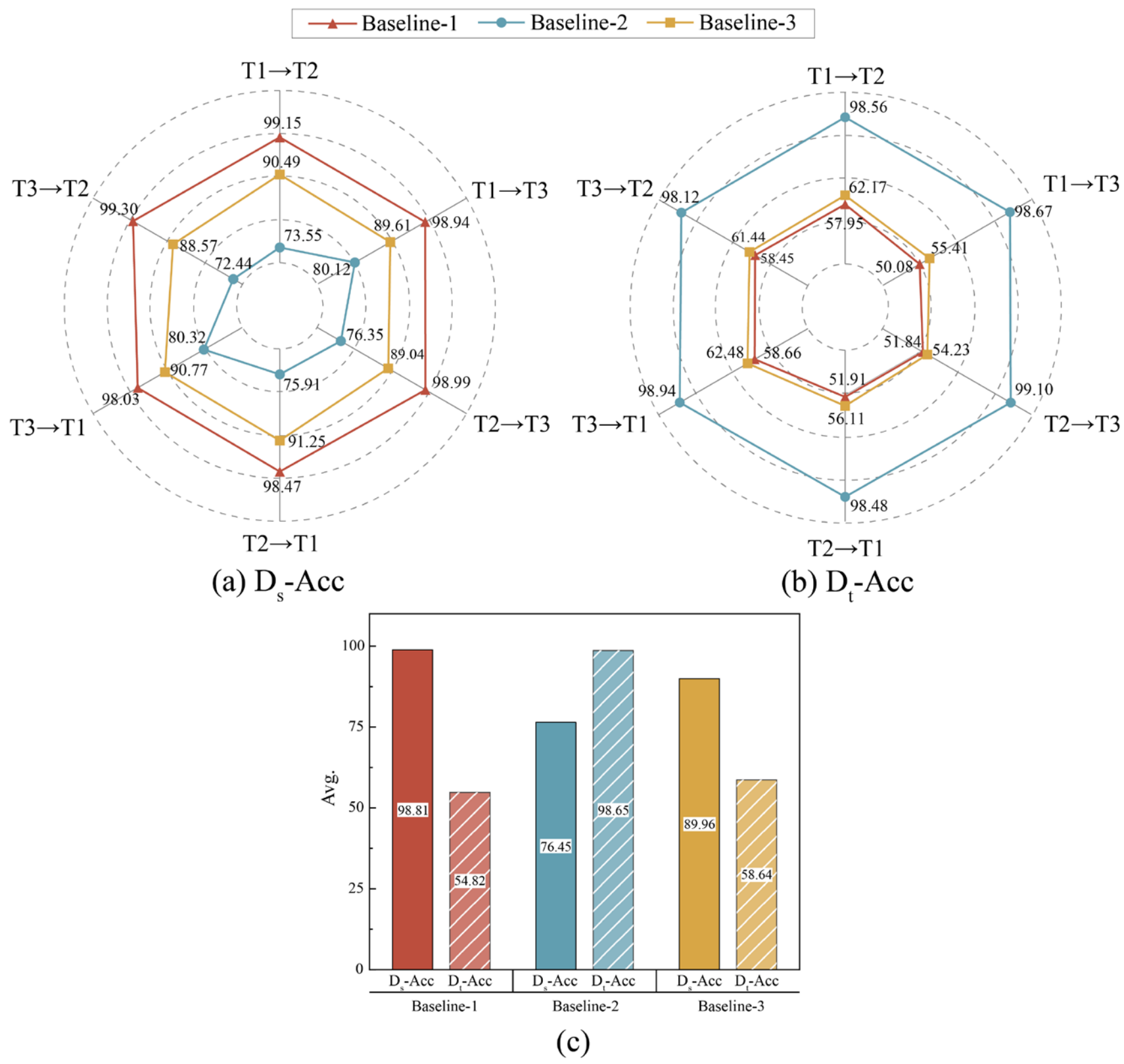

**Fig. 11** The baseline's accuracy results on LSPS dataset. (a) The accuracy results on $D_S$. (b) The accuracy results on $D_t$.(c) The average accuracy results on LSPS dataset. (For an explanation of the color references in the legend of this figure, please refer to the online version of this paper.)

The experimental results indicate that the three baseline networks establish the performance boundaries of three transfer learning strategies: zero-shot adaptation (Baseline-1), full fine-tuning (Baseline-2), and selective fine-tuning (Baseline-3). Baseline-1 exhibits high accuracy on the source domain dataset; however, its accuracy significantly decreases when evaluated on the target domain. This highlights substantial differences in the feature distributions between the source and target domains, suggesting that a single model is insufficient to effectively encompass the feature spaces of both domains. Further analysis reveals a positive correlation between the performance differences among subsets of the target domain (e.g., T3 → T1: 58.66 %) and the degree of overlap in their respective feature spaces, thereby confirming the critical role of feature space overlap in cross-domain generalization.

Baseline-2 employs a pretraining followed by full-parameter fine-tuning strategy, which significantly enhances target domain accuracy (TIGFH: +43.83 %, LSPS: +47.23 %) but leads to a degradation of source domain performance (TIGFH: -22.36 %, LSPS: -24.41 %). This suggests that the model experiences difficulty in retaining knowledge acquired from the source domain while adapting to the target domain, a phenomenon referred to as catastrophic forgetting [34]. This highlights an inherent conflict between straightforward parameter-sharing approaches and the goal of improving cross-domain generalization capabilities.

To mitigate catastrophic forgetting, the Baseline-3 network restricts training to only the classification layer and the final convolutional layer on the target domain [35]. As shown in **Fig. 11**, this approach achieves a moderate reduction in the model's forgetting of source domain knowledge (TIGFH: -8.85 %, LSPS: -8.86 %) but offers limited improvement in target domain accuracy (TIGFH: +3.82 %, LSPS: +4.47 %). Furthermore, it fails to significantly enhance generalization capabilities. This approach requires manual selection of layers for fine-tuning, which limits the model's adaptability to complex domain shifts and overlooks the suboptimal allocation of layers with respect to the significance of specific domain features. Consequently, it demonstrates insufficient robustness when confronted with substantial domain shifts across different processes. These limitations highlight the inadequacy of this strategy in scenarios where significant discrepancies exist between feature spaces of different domains, thereby motivating our investigation into UDA methods that incorporate dynamic parameter optimization and learnable domain alignment mechanisms.

In summary, when a significant feature disparity exists between the source and target domains, basic transfer learning strategies fail to achieve satisfactory generalization. Although training with both source and target domain data is a feasible solution, it frequently encounters two major challenges in practical applications: high cost of data annotation and low training efficiency. The limitations of existing methods highlight the importance of simultaneously addressing the needs for "no target domain annotations" and "mitigating inter-domain feature discrepancies." UDA methods, by modeling the mapping of feature spaces between the source and target domains, can effectively utilize unlabeled target domain data, thereby offering a novel technical approach to resolve the aforementioned challenges. This not only reduces the time and labor expenses associated with annotating new data but also facilitates the efficient transfer of acquired knowledge from one domain to another similar domain.

### 4.2 Same-process transfer experiment

To ensure a fair and rigorous comparison in the subsequent experiments, all baseline methods utilize the same fundamental setup as our model: they are trained on identical hardware and evaluated on the same datasets using a uniform accuracy metric. Crucially, to assess each method's performance, we adopted the optimal hyperparameters for each baseline (e.g., MSTN, FixBi) as reported in their original publications. Our own model's hyperparameters (SGD optimizer, LR of 0.002, OneCycleLR, batch size 64) are those we found to be optimal for our proposed model.

**Tab. 5** and **Tab. 6** present the results of same-process transfer experiments for the TIG and laser welding datasets, respectively. The accuracy metrics were evaluated on the target domain. The table reports the accuracy results obtained from five times using different random seeds. **Fig. 12** illustrates the training curves corresponding to the two subtasks.

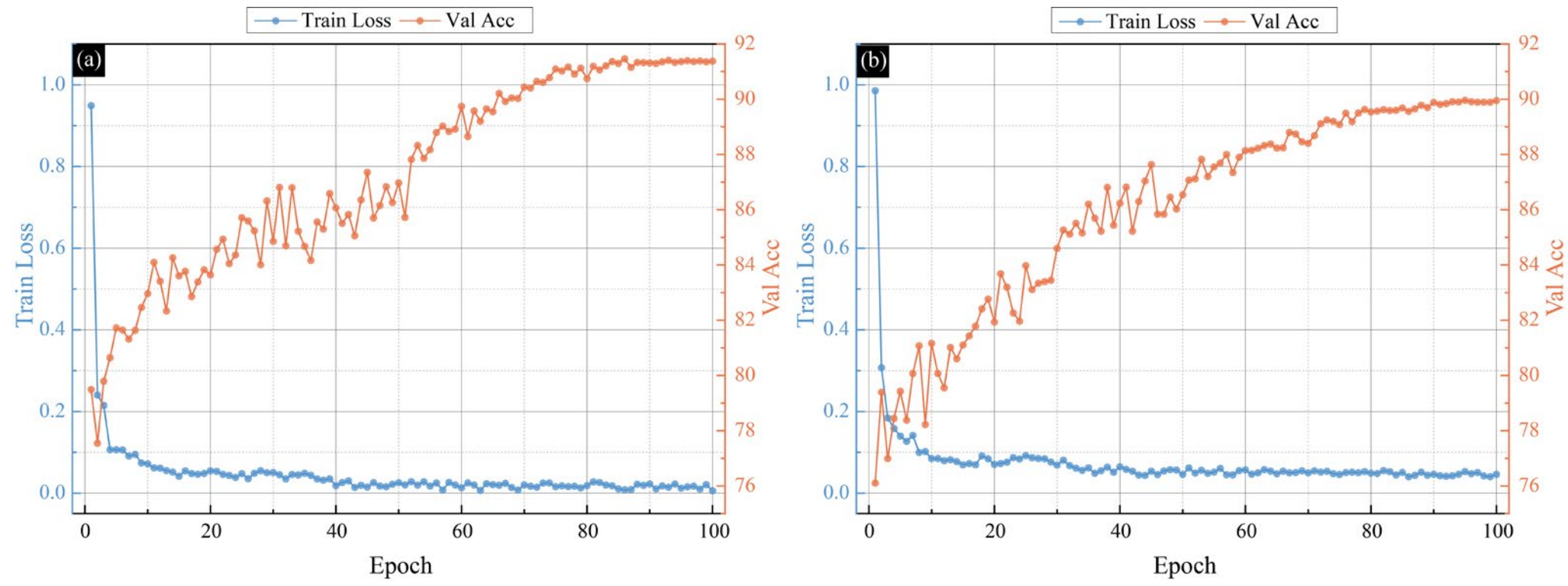

**Fig. 12** (a) The training curve for the T1 → T2 task; (b) The training curve for the L3 → L1 task

In the same-process transfer task of the TIGFH dataset, the method proposed in this study exhibits significant advantages, achieving an average accuracy of 90.65 %, which represents a 4.24 % improvement over the second-best method FixBi (86.41 %). Additionally, it exhibits a lower standard deviation compared to other approaches. This demonstrates that the GSDE strategy, by modeling the associated feature space of welding process parameters and CMOS imaging angles, substantially enhances the model's adaptability to variations within the same-domains.

**Tab. 5** Accuracy results on TIGFH dataset. Best results are displayed in bold and the runner-up results are underlined, utilizing five distinct random seeds.

| Method | T1 → T2 | T1 → T3 | T2 → T3 | T2 → T1 | T3 → T1 | T3 → T2 | $Avg.\uparrow$ | $\sigma\downarrow$ |
|---|---|---|---|---|---|---|---|---|
| Baseline-1 | 57.95 | 50.08 | 51.84 | 51.91 | 58.66 | 58.45 | 54.82 | 3.59 |
| MSTN [26] | 82.82±0.98 | 78.34±0.87 | 81.87±0.83 | 82.64±0.93 | 79.37±0.97 | 78.07±1.00 | 80.52 | 1.99 |
| CDAN+E [25] | 83.67±0.78 | 83.58±0.81 | 83.21±0.75 | 80.64±0.79 | 79.02±0.81 | 79.70±0.92 | 81.63 | 1.99 |
| GVB-GD [31] | 80.94±0.72 | 85.56±0.76 | 84.27±0.68 | 83.63±0.72 | 85.47±0.63 | 82.71±0.80 | 83.76 | 1.61 |
| DCAN [36] | 83.30±0.48 | 85.84±0.54 | 85.68±0.47 | 85.51±0.30 | 84.83±0.34 | 87.84±0.42 | 85.50 | 1.35 |
| FixBi [37] | 84.82±0.55 | 85.40±0.52 | **90.42±0.49** | 85.26±0.57 | 83.80±0.54 | <u>88.74±0.59</u> | 86.41 | 2.35 |
| Ours | <u>88.84±0.44</u> | <u>89.54±0.35</u> | 86.40±0.43 | <u>87.58±0.41</u> | <u>87.54±0.39</u> | 87.26±0.42 | <u>87.86</u> | <u>1.03</u> |
| Ours+GSDE | **90.87±0.45** | **91.38±0.36** | <u>89.57±0.49</u> | **90.55±0.40** | **91.62±0.38** | **89.96±0.44** | **90.65** | **0.72** |

In the same-process transfer task of the LSPS dataset, the method proposed in this paper achieves a leading average accuracy of 90.72 %, with the lowest standard deviation compared to all other approaches. This method demonstrates robust performance across various complex parameter variations, including camera placement, workpiece thickness, point-ring laser power, and external light source differences. Notably, in the L2→L1 transfer task, the method attains an accuracy of 91.34 %, which represents a significant 5.93 % improvement over the FixBi method. These results indicate that the GSDE strategy effectively mitigates feature distribution shift during transfer process.

**Tab. 6** Accuracy results on LSPS dataset. Best results are displayed in bold and the runner-up results are underlined, utilizing five distinct random seeds.

| Method | L1 → L2 | L1 → L3 | L2 → L3 | L2 → L1 | L3 → L1 | L3 → L2 | $Avg.\uparrow$ | $\sigma\downarrow$ |
|---|---|---|---|---|---|---|---|---|
| Baseline-1 | 50.12 | 49.20 | 48.12 | 49.50 | 53.46 | 60.70 | 51.85 | 4.29 |

| MSTN [26] | 81.64±0.89 | 83.80±0.92 | 82.21±0.75 | 82.63±1.06 | 82.70±0.87 | 79.53±0.96 | 82.10 | 1.31 |
|---|---|---|---|---|---|---|---|---|
| CDAN+E[25] | 84.30±0.78 | 87.17±0.62 | 78.45±0.81 | 83.50±0.74 | 80.18±0.59 | 81.24±0.61 | 82.47 | 2.87 |
| GVB-GD[31] | 83.88±0.48 | 86.91±0.43 | 85.52±0.31 | 84.83±0.57 | 85.07±0.43 | 83.50±0.45 | 84.95 | 1.11 |
| DCAN [36] | 84.93±0.35 | 84.41±0.34 | 85.57±0.31 | 87.04±0.37 | 84.67±0.38 | 86.80±0.30 | 85.57 | 1.02 |
| FixBi [37] | 88.82±0.48 | 89.85±0.46 | 86.14±0.44 | 85.41±0.51 | 85.84±0.42 | 85.53±0.49 | 86.93 | 1.74 |
| Ours | 87.89±0.39 | 91.05±0.33 | 87.89±0.34 | 87.17±0.41 | 86.38±0.39 | 87.59±0.42 | 87.99 | 1.46 |
| Ours+GSDE | **90.02±0.36** | **92.64±0.32** | **89.73±0.30** | **91.34±0.29** | **89.87±0.34** | **90.74±0.31** | **90.72** | **0.86** |

### 4.3 Cross-process transfer experiments

In the TIGFH→LSPS cross-process transfer task (**Tab. 7**), the method proposed in this study demonstrated significantly superior performance, achieving an average accuracy of 80.48 %, This result confirms that the GSDE strategy effectively captures cross-process correlations between TIG welding and laser welding. In the reverse transfer task, LSPS→TIGFH (**Tab. 8**), the proposed method also maintained optimal performance, with an average accuracy of 81.36 % and a standard deviation lower than all other compared methods. This indicates that the GSDE strategy successfully mitigates high-order semantic mismatch between different welding processes by leveraging pseudo-source samples. Further analysis of the bidirectional transfer experiment revealed a minimal performance difference of only 0.88 % (80.48 % vs. 81.36 %) between TIGFH → LSPS and LSPS → TIGFH, which is considerably smaller than that of FixBi (1.65 %) and GVB-GD (2.90 %). This suggests that GSDE achieves robust bidirectional transfer capability through process-agnostic feature representation. **Fig. 13** illustrates the training curves corresponding to the two subtasks.

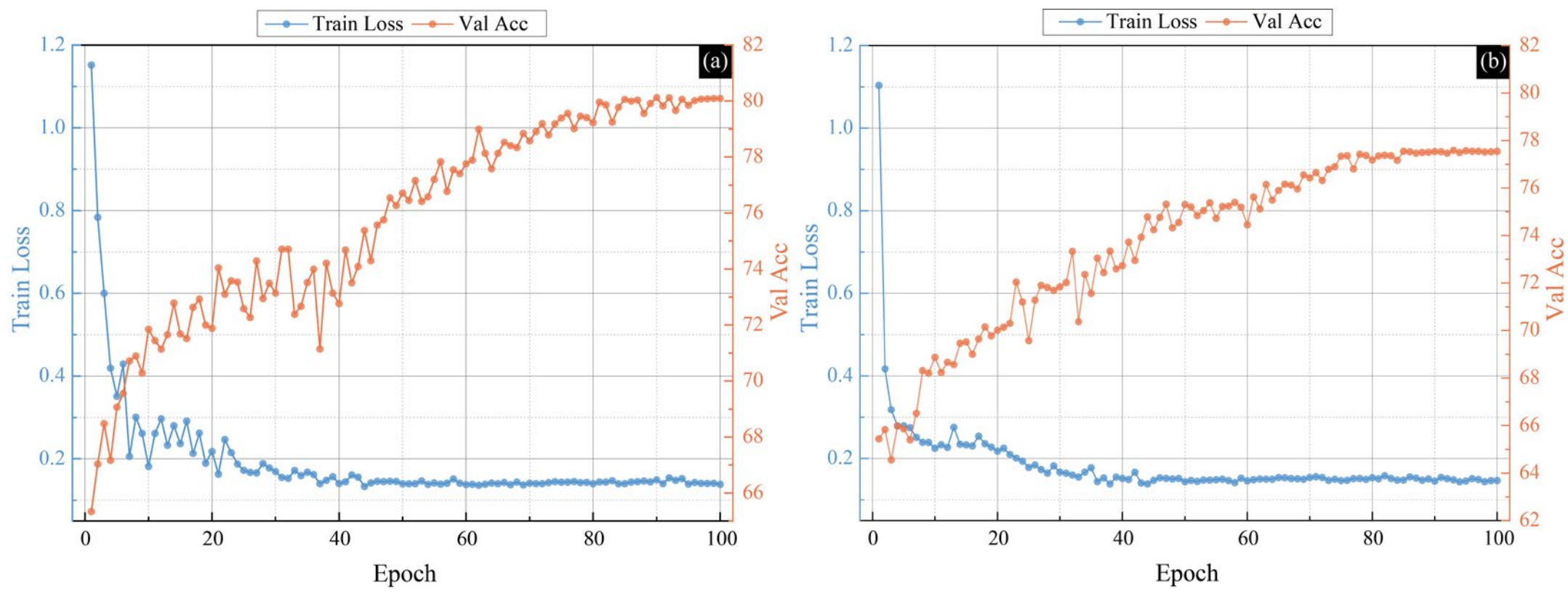

**Fig. 13** (a) The training curve for the T3→L3 task; (b) The training curve for the L1→T2 task

The experimental results indicate that traditional transfer learning methods struggle to align feature spaces effectively in cross-process scenarios. Their performance is limited by the high-order semantic mismatch between process parameters. The model proposed in this study constructs an intermediate semantic space across processes via adversarial training, thereby bridging the morphological disparity in molten pools between TIG welding and laser welding. The results substantiate the efficacy of GSDE's core design: 1) The adversarial adaptation method aligns the feature spaces of melt pool morphology variations across different welding processes by optimizing the statistical distance between the source and target domains. In **Tab. 8**, our method's standard deviation (σ =1.82) is significantly lower than FixBi's (σ = 2.71), indicating that the pseudo-source data guidance mechanism effectively mitigates the distributional shift of semantic information in high-dimensional space; 2) The entropy minimization strategy maintains classification consistency under fluctuations in process parameters by driving the decision boundary away from the data; 3) The dynamic construction of the pseudo-source domain based on high-confidence pseudo-labels further bridges the high-order semantic differences among various welding process parameters, resulting in symmetrical bidirectional transfer performance. It should be noted that although the current cross-process accuracy still lags behind within-process transfer, GSDE systematically addresses feature shift and boundary ambiguity issues through a three-tier architecture (global alignment → boundary optimization → fine-grained correction), thereby providing a viable technical framework for industrial transfer learning. For future research, two potential optimization directions are proposed: a) employing more advanced domain alignment strategies to more effectively extract domain-invariant features from both source and target domains, thereby ensuring precise alignment; b) integrating the differences in physical mechanisms between TIG welding and Laser welding into the pseudo-label generation process to minimize the likelihood of incorrect pseudo-label generation, thereby guiding the domain alignment process in UDA.

**Tab. 7** Accuracy results from TIGFH to LSPS dataset. Best results are displayed in bold and the runner-up results are underlined, utilizing five distinct random seeds.

| Method | T1→L1 | T1→L2 | T1→L3 | T2→L1 | T2→L2 | T2→L3 | T3→L1 | T3→L2 | T3→L3 | Avg. ↑ | σ ↓ |
|---|---|---|---|---|---|---|---|---|---|---|---|
| Baseline-1 | 34.84 | 30.82 | 37.04 | 38.7 | 43.51 | 30.17 | 48.43 | 38.91 | 31.43 | 37.09 | 5.77 |
| MSTN[26] | 55.90 ±1.27 | 62.82 ±1.30 | 57.21 ±1.25 | 58.53 ±1.14 | 57.56 ±1.16 | 62.64 ±1.04 | 68.32 ±1.10 | 65.05 ±1.05 | 66.43 ±1.16 | 61.61 | 4.22 |
| CDAN+E[25] | 62.15 ±0.75 | 67.15 ±0.66 | 73.24 ±0.82 | 73.73 ±0.69 | 64.35 ±0.87 | 65.15 ±0.85 | 67.15 ±0.60 | 69.76 ±0.76 | 67.40 ±0.84 | 67.79 | 3.66 |
| GVB-GD[31] | 65.08 ±0.84 | 72.53 ±0.57 | 74.31 ±0.77 | 75.12 ±0.68 | 73.11 ±1.44 | 70.07 ±0.94 | 74.91 ±0.54 | 67.64 ±0.57 | 72.21 ±0.84 | 71.66 | 3.24 |
| DCAN[36] | 72.40 ±1.56 | 73.28 ±1.54 | 74.57 ±1.98 | 75.46 ±1.42 | 74.39 ±1.15 | 70.69 ±1.49 | 72.71 ±1.47 | 76.12 ±2.21 | 75.62 ±1.56 | 73.91 | 1.67 |
| FixBi[37] | 73.82 ±0.75 | 73.02 ±0.84 | 75.45 ±0.77 | 77.12 ±0.84 | 72.32 ±0.83 | 74.35 ±0.75 | 75.14 ±1.68 | 74.24 ±0.57 | 77.08 ±0.55 | 74.72 | 1.56 |
| Ours | 76.72 ±0.51 | 76.04 ±0.48 | 78.11 ±0.47 | 79.32 ±0.47 | 78.47 ±0.52 | 76.99 ±0.55 | 78.00 ±0.37 | 75.45 ±0.39 | 78.12 ±0.37 | 77.47 | 1.17 |
| Ours+GSDE | **80.93** ±0.46 | **79.13** ±0.43 | **79.94** ±0.47 | **81.32** ±0.44 | **81.65** ±0.45 | **81.06** ±0.32 | **81.39** ±0.35 | **78.49** ±0.36 | **80.12** ±0.38 | **80.48** | **0.99** |

**Tab. 8** Accuracy results from LSPS to TIGFH dataset. Best results are displayed in bold and the runner-up results are underlined, utilizing five distinct random seeds.

| Method | L1→T1 | L1→T2 | L1→T3 | L2→T1 | L2→T2 | L2→T3 | L3→T1 | L3→T2 | L3→T3 | Avg.↑ | σ↓ |
|---|---|---|---|---|---|---|---|---|---|---|---|
| Baseline-1 | 36.51 | 31.28 | 34.95 | 37.48 | 41.98 | 31.25 | 38.51 | 46.35 | 41.23 | 37.73 | 4.69 |
| MSTN[26] | 61.72 ±1.22 | 64.57 ±1.10 | 56.76 ±1.05 | 71.45 ±1.06 | 71.64 ±1.18 | 70.32 ±1.11 | 71.48 ±1.45 | 72.15 ±1.15 | 70.20 ±0.89 | 67.81 | 5.18 |
| CDAN+E[25] | 69.22 ±1.02 | 71.84 ±0.97 | 73.47 ±1.01 | 74.02 ±1.15 | 75.32 ±1.10 | 76.15 ±1.08 | 75.55 ±1.11 | 74.74 ±0.98 | 70.52 ±1.02 | 73.42 | 2.26 |
| GVB-GD[31] | 75.00 ±0.96 | 75.34 ±0.90 | 78.47 ±1.73 | 75.48 ±0.89 | 78.74 ±1.76 | 73.53 ±0.83 | 74.49 ±0.95 | 76.05 ±0.86 | 71.21 ±0.82 | 75.36 | 2.18 |
| DCAN[36] | 74.65 ±0.86 | 75.40 ±0.79 | 79.28 ±1.86 | 75.07 ±0.71 | 78.09 ±0.82 | 77.03 ±0.87 | 74.67 ±0.95 | 78.21 ±0.88 | 73.24 ±0.70 | 76.18 | 1.92 |
| FixBi[37] | 75.54 ±0.74 | 73.67 ±0.88 | 76.14 ±0.74 | 76.00 ±0.74 | 79.19 ±0.74 | 78.13 ±0.65 | 77.09 ±1.98 | 79.54 ±0.74 | 72.11 ±0.74 | 76.37 | 2.31 |
| Ours | 77.15 ±0.57 | 75.62 ±0.44 | 80.89 ±0.73 | 76.58 ±0.49 | 81.24 ±0.52 | 78.67 ±0.49 | 77.18 ±0.55 | 80.55 ±0.58 | 76.02 ±0.61 | 78.24 | 2.03 |
| Ours+GSDE | **79.42 ±0.49** | **77.84 ±0.43** | **82.92 ±0.46** | **80.25 ±0.32** | **84.52 ±0.41** | **81.92 ±0.45** | **79.75 ±0.33** | **83.69 ±0.49** | **79.94 ±0.47** | **81.13** | **1.82** |

## 4.4 Ablation Experiment

### 4.4.1 Impact of pseudo-source proportion

This subsection analyzes the model's sensitivity to the proportion of pseudo-source samples. We conducted experiments with fixed proportions by selecting the top 10 %, 20 %, and 50 % of target samples based on prediction confidence. As shown in **Fig. 14**, which plots the mean cross-process accuracy, the results reveal a critical trade-off. An overly small proportion provides insufficient guidance for domain alignment, leading to only marginal accuracy gains (e.g., from a 74.43 % to 76.68 %). Conversely, an excessively large proportion incorporates noisy, low-confidence samples, whose erroneous pseudo-labels misguide the alignment process and degrade performance (accuracy drops to 68.09 %). This analysis validates our proposed strategy of progressively increasing the pseudo-source proportion, as it dynamically balances effective guidance with noise injection.

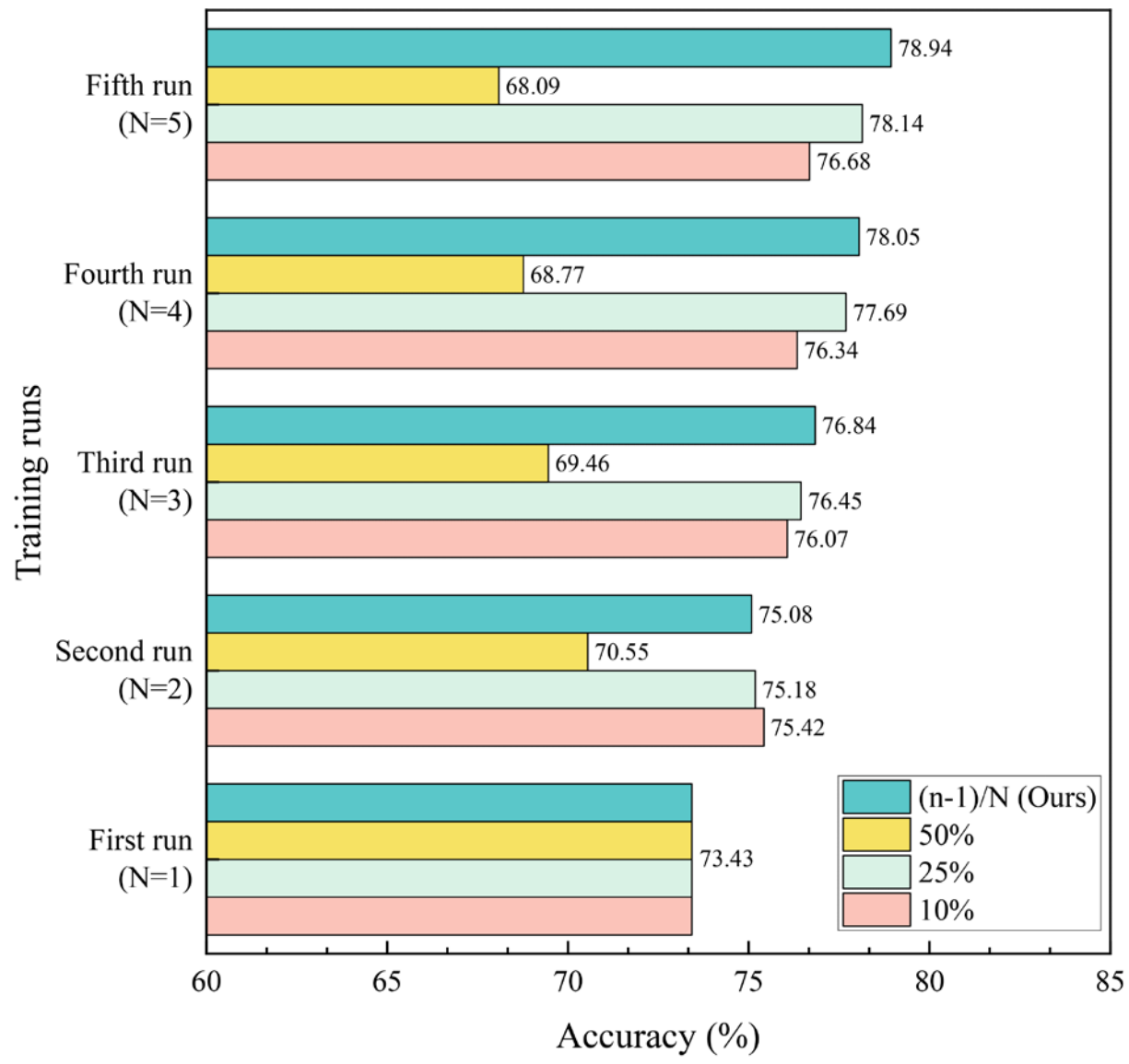

**Fig. 14** Comparison of model accuracy across five training epochs under varying proportions of pseudo-labels.

### 4.4.2 Number of max run $N$

The algorithm's performance was evaluated across multiple datasets for varying maximum iteration counts $N$. In this ablation experiment, the number of bottlenecks $k$ was fixed at 1. As illustrated in **Fig. 14(a)**, when $N \leq 5$, the algorithm's accuracy increases sharply with increasing $N$; when $N > 5$, although accuracy continues to improve, the rate of enhancement significantly diminishes. This suggests that larger $N$ values may yield better results. However, given that the computational cost scales linearly with $N$, $N = 5$ was selected in this study to achieve an optimal balance between computational efficiency and accuracy.

4.4.3 Number of bottlenecks $k$

This study employs varying numbers of bottleneck to evaluate the model. In this ablation experiment, the maximum number of iterations, $N$, was fixed at 1. **Fig. 14(b)** illustrates the effect of the number of bottlenecks on model accuracy. The results indicate that as the number of bottlenecks increases, the model's accuracy improves accordingly. When $k = 4$, the model reaches its peak average accuracy of 81.68 %. However, as $k$ continues to increases beyond this point, the accuracy begins to decline. Consequently, the number of bottlenecks, $k$, is set to 4 for optimal performance.

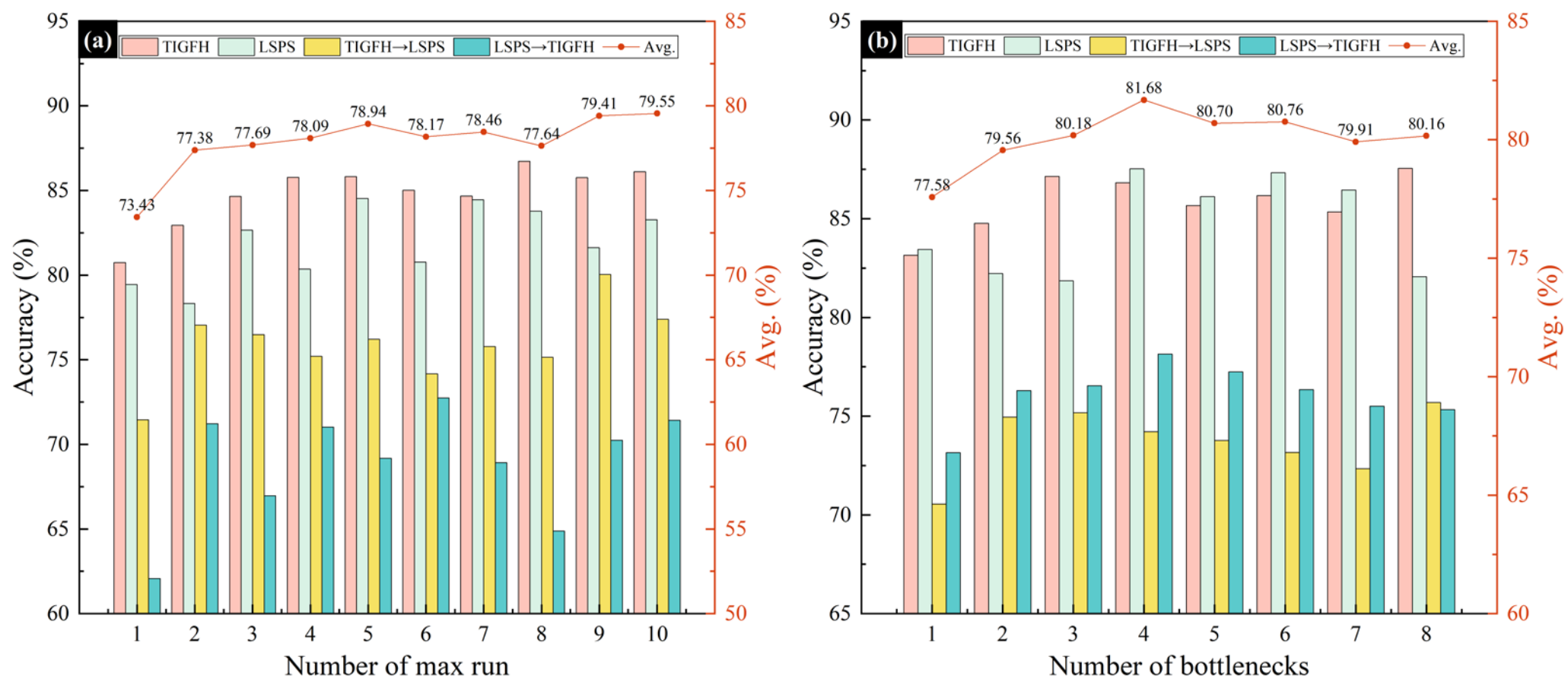

**Fig. 15** (a) Accuracy over number of max run; (b) Accuracy over number of bottlenecks.

4.4.4 Reinitialization and source domain expansion:

Initially, the impact of re-initializing network weights at each run on model accuracy was investigated. Specifically, the ablation study retained the weights trained in the previous run rather than re-initializing them. Additionally, the advantage of expanding the source domain over relying solely on pseudo-labels for classification loss was explored. In this context, pseudo-labels were utilized with only the $L_{CLAS}$ component without incorporating the other three losses. The baseline exhibited stable performance on the same-process transfer task but demonstrated a significant decrease in cross-process transfer task (TIGFH → LSPS: 76.41 %, LSPS → TIGFH: 78.31 %). This result indicates that a simple classification loss struggles to capture cross-welding-process semantic correlations. Upon introducing weight re-initialization (Init), only marginal improvements were observed in both same-process transfer task accuracy (TIGFH: +0.44 %, LSPS: +0.67 %) and cross-process transfer task performance (TIGFH: +0.51 %, LSPS: +0.71 %). These findings suggest that while historical training parameters can perpetuate same-process feature representation, they are insufficient to bridge the cross-process semantic gap. Optimal bidirectional transfer task performance was achieved by combining weight re-initialization with source domain expansion, underscoring the value of pseudo-source data for adversarial loss in guiding effective domain alignment. (**Tab. 9**)

**Tab. 9** Impact of re-initialization and source domain expansion on model accuracy. Best results are displayed in bold and the runner-up results are underlined.

| | TIGFH | LSPS | TIGFH → LSPS | LSPS → TIGFH |
|---|---|---|---|---|
| Baseline | 87.97 | 87.97 | 76.41 | 78.31 |
| + Init | 88.41 (+0.44) | 88.64 (+0.67) | 76.92 (+0.51) | 79.02 (+0.71) |
| + expansion | <u>89.56 (+1.15)</u> | <u>89.81 (+1.17)</u> | <u>79.53 (+2.61)</u> | <u>83.37 (+1.06)</u> |
| + Loss | **90.07 (+0.51)** | **90.57 (+0.76)** | **80.08 (+0.55)** | **80.54 (+0.46)** |

4.4.5 Contribution of each adaptation loss:

As shown in **Tab. 10**, this study conducts a quantitative analysis of the contribution of the three loss functions to the performance of the UDA model. The results indicate that $L_{MS}$ and $L_{SS}$ exert distinct influences on improving model accuracy, with $L_{MS}$ demonstrating a more significant contribution. When $L_{ADV}$ is utilized independently, it achieves an accuracy of 74.12 % in the TIGFH → LSPS transfer task, where its calibration of the inter-domain category center shift is likely the primary factor contributing to the accuracy improvement. After implementing $L_{SS}$, an additional performance enhancement of 1.44 % is observed in the reverse transfer task, surpassing the 0.23 % improvement noted in the forward transfer. This advantage may stem from performing multiple augmentations on unlabeled data and calculating their average predictions. Concurrently, MixUp integrates labeled and unlabeled data through linear interpolation, further enhancing the model's generalization capacity and robustness.

**Tab. 10** Contribution of each adaptation loss. Best results are displayed in bold and the runner-up results are underlined.

| | TIGFH | LSPS | TIGFH → LSPS | LSPS → TIGFH |
|---|---|---|---|---|
| $L_{ADV}$ | 86.15 | 84.00 | 74.12 | 74.63 |
| $L_{ADV} + L_{MS}$ | <u>89.84 (+3.69)</u> | <u>89.13 (+5.13)</u> | <u>79.28 (+5.16)</u> | <u>79.81 (+5.18)</u> |
| $L_{ADV} + L_{MS} + L_{SS}$ | **90.07 (+0.23)** | **90.57 (+1.44)** | **80.08 (+0.80)** | **80.54 (+0.73)** |

**4.5 UMAP visualization**

To validate the effectiveness of the UDA model in cross-domain transfer tasks, this study employs the UMAP algorithm [38] for feature space visualization. UMAP has been demonstrated to offer significant advantages over other topology-based dimensionality reduction techniques, such as PCA [39], Isomap [40], and t-SNE [41], in terms of visualization quality, preservation of the global structure of data, and computational efficiency. Its primary workflow consists of two key steps: first, constructing a high-dimensional graph representation of the data to represent 'fuzzy simplicial complexes'; second, minimizing the cross-entropy between two fuzzy sets that share the same underlying elements (data points), thereby optimizing low-dimensional vectors to approximate the high-dimensional graph representation as closely as possible. As illustrated in **Fig. 15**, this study performed feature visualization analysis for both same-process and cross-process transfer tasks using the TIGFH and LSPS datasets (sample size per class: $n = 600$). In the figures, yellow and red data points represent the penetration state in the source and target domains, respectively, while blue and green data points represent the non-penetration state in the source and target domains, respectively. The black dashed line represents the potential decision boundary. It can be seen from the figures that the model's approach to modeling the data distribution varies across different transfer tasks. In the same-process transfer tasks (**Fig. 15(a)** and **Fig. 15(b)**), the features show several positive characteristics: samples are compactly distributed, intra-class distributions are similar, and inter-class boundaries are clear. Furthermore, samples are distant from the decision boundary, and the features of both domains are well-aligned. In contrast, in the cross-process transfer tasks (**Fig. 15(c)** and **Fig. 15(d)**), compared to the same-process transfer tasks, feature samples from different classes exhibit a mixing phenomenon. The features are not completely aligned, intra-class sample distributions are dispel,rsed, and misclassified feature points tend to cluster into isolated groups. This situation may stem from inaccuracy in the soft labels for pseudo-source domain data, leading to some feature samples being incorrectly classified. Simultaneously, a significant number of feature samples cluster near the decision boundary, resulting in a notable difference in model accuracy compared to the same-process transfer tasks. This reflects the domain shift caused by differences in the opto-thermal-electrical coupling mechanisms of various welding processes. This phenomenon highlights the fundamental impact of physical mechanism differences between processes on the feature space distribution, providing crucial insights for improving UDA methods.

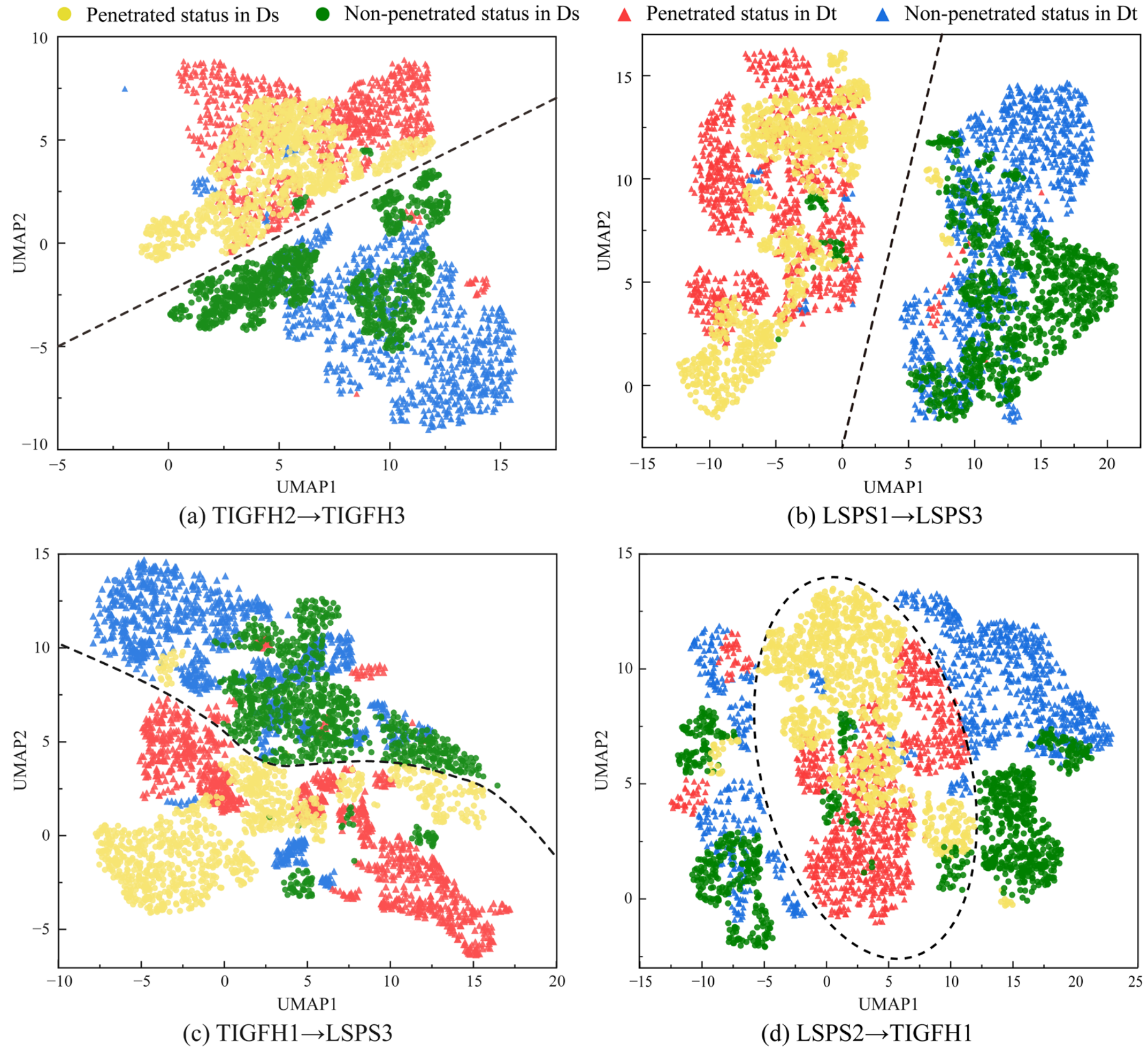

**Fig. 16** Model visualization using UMAP. (a) and (b) illustrate the visualization results of transfer tasks within the same-process; (c) and (d) illustrate the visualization results of transfer tasks within cross-process. (For an explanation of the color references in the legend of this figure, please refer to the online version of this paper.)

## 5. Conclusion

This study systematically investigated the feasibility of UDA transfer learning in cross-process scenarios, and demonstrated the effectiveness of the proposed method in both TIG welding (TIGFH dataset) and laser welding (LSPS dataset) contexts. The research focused on a comparative analysis of same-process and cross-process transfer learning methods, leading to the following key conclusions:

1） A novel cross-process UDA model was designed to overcome the significant domain shift caused by disparate physical mechanisms in TIG and laser welding. The model architecture leverages a ResNet-50 backbone and introduces a gradual source domain expansion (GSDE) strategy. This algorithm progressively incorporates high-confidence pseudo-labeled target samples into an expanded source domain, providing crucial guidance for feature alignment. The training was optimized through a composite loss function that integrated adversarial, semantic, and semi-supervised losses, enabling the model to learn domain-invariant features while preserving class-specific structures.

2） In same-process transfer tasks, the proposed method achieved an average accuracy of 90.65 % on the TIGFH dataset, representing a 4.41% improvement over the second-best method FixBi (86.85 %). On the LSPS dataset, it reached an average accuracy of 90.72 %, significantly outperforming other methods. In cross-process transfer tasks, the method achieved 80.48 % accuracy in the TIGFH → LSPS task and 81.13 % in the LSPS → TIGFH task, with standard deviations of 0.99 and 1.82 respectively, substantially lower than existing methods. Notably, the performance difference between bidirectional transfer tasks was only 0.65 % (80.48 % vs. 81.13 %), considerably smaller than FixBi (2.01 %) and GVB-GD (3.70 %), indicating that GSDE achieved robust bidirectional transfer capability through process-agnostic feature representation.

3） The UMAP visualization results confirmed that the UDA model effectively bridges the topological discrepancies within cross-process feature spaces through pseudo-source sample generation and a progressive feature alignment strategy, while maintaining relatively distinct class boundaries. By eliminating the need for target domain annotations, our approach significantly reduced the cost and time associated with model training, providing a viable solution for the implementation of adaptive and intelligent welding quality control systems in industrial environments.

# Declarations

**Ethics approval** Not applicable.

**Consent to participate** The authors declare their consent to participate.

**Consent for publication** The authors declare their consent for publication.

**Conflict of interest** The authors declare no competing interests.

# Funding

The authors gratefully acknowledge the financial support from the National Key R&D Program of China (No. 2023YFB3407800), National Natural Science Foundation of China (No. U2141213).